%% file: main.tex
\documentclass{article}

\usepackage{microtype}
\usepackage{graphicx}
\usepackage{booktabs} 

\usepackage{hyperref}

\usepackage[accepted]{icml2024}

\usepackage{amsmath}
\usepackage{amssymb}
\usepackage{mathtools}
\usepackage{amsthm}
\usepackage{listings}
\usepackage{afterpage}
\usepackage{multirow}
\usepackage{makecell}
\usepackage{ulem}
\usepackage{tcolorbox}
\usepackage{subfig}
\usepackage{caption}
\usepackage{titlesec}
\titlespacing*{\section}{0pt}{0.4ex}{0.4ex}  
\titlespacing*{\subsection}{0pt}{0.3ex}{0.3ex}
\titlespacing*{\subsubsection}{0pt}{0.3ex}{0.3ex}
\usepackage{xspace}
\newcommand{\leaner}{\textsc{Leaner}\xspace}
\newcommand{\glue}{\textsc{Glue}\xspace}
\newcommand{\bert}{\textsc{Bert}\xspace}
\newcommand{\llama}{\textsc{Llama}\xspace}

\newcommand{\xlnet}{\textsc{Xlnet}\xspace}
\newcommand{\mamba}{\textsc{Mamba}\xspace}
\newcommand{\gpt}{\textsc{Gpt}\xspace}
\newcommand{\explm}{\textsc{Exp1}\xspace}
\newcommand{\expinstruct}{\textsc{Exp2}\xspace}
\newcommand{\expcurriculum}{\textsc{Exp3}\xspace}
\newcommand{\itersentlen}{\textsc{Iter. SentLen}\xspace}
\newcommand{\iterselfloss}{\textsc{Iter. Self-Loss}\xspace}
\newcommand{\iterlmloss}{\textsc{Iter. LM-Loss}\xspace}
\newcommand{\fullrepeated}{\textsc{Full Repeated}\xspace}
\newcommand{\fullrandom}{\textsc{Full Random}\xspace}
\newcommand{\iterrandom}{\textsc{Iter. Random}\xspace}
\usepackage{tikz}

\newcommand{\colorfulcircle}[1]{%
    \begin{tikzpicture}
        \fill[#1] (0,0) circle(0.4*\the\ht\strutbox);  
    \end{tikzpicture}%
}

\usepackage{xcolor}
\definecolor{airforceblue}{rgb}{0.36, 0.54, 0.66}
\definecolor{bluegray}{rgb}{0.4, 0.6, 0.8}
\definecolor{bleudefrance}{rgb}{0.19, 0.55, 0.91}
\hypersetup{colorlinks,linkcolor={airforceblue},citecolor={bleudefrance},urlcolor={bluegray}}

\definecolor{new_idea}{rgb}{0.51, 0.63, 0.84}
\definecolor{warning}{rgb}{0.72, 0, 0}
\definecolor{leaner}{rgb}{0.180, 0.325, 0.490}
\definecolor{original}{rgb}{0.4, 0.4, 0.4}
\definecolor{lightbrown}{rgb}{0.949, 0.710, 0.420}
\definecolor{mediumbrown}{rgb}{0.749, 0.522, 0.388}
\definecolor{darkbrown}{rgb}{0.549, 0.376, 0.337}
\definecolor{color-empty}{HTML}{A3A3A3}
\definecolor{color-full-repeated}{HTML}{BF8563}
\definecolor{color-full-random}{HTML}{8C6056}
\definecolor{color-iter-random}{HTML}{F2B56B}
\definecolor{color-iter-sentlen}{HTML}{77BDD9}
\definecolor{color-iter-selfloss}{HTML}{245B8C}
\definecolor{color-iter-lmloss}{HTML}{D93232}

\newcommand{\best}[1]{\textcolor{mediumbrown}{\textit{\textbf{#1}}}}
\newcommand{\better}[1]{\textcolor{darkbrown}{\textbf{#1}}}
\newcommand{\improvement}[1]{\textcolor{mediumbrown}{\textit{\textbf{(+#1)}}}}

\usepackage[capitalize,noabbrev]{cleveref}

\theoremstyle{plain}

\theoremstyle{definition}

\theoremstyle{remark}

\NewDocumentCommand{\cheng}{ mO{} }{\textcolor{blue}{\textsuperscript{\textit{Cheng}}\textsf{\textbf{\small[#1]}}}}

\usepackage[textsize=tiny]{todonotes}

\icmltitlerunning{TinyHelen's First Curriculum: Training and Evaluating Tiny Language Models in a Simpler Language Environment}

\begin{document}
\twocolumn[
\icmltitle{TinyHelen's First Curriculum: Training and Evaluating\\Tiny Language Models in a Simpler Language Environment}


\icmlsetsymbol{equal}{*}

\begin{icmlauthorlist}
\icmlauthor{Ke Yang}{uiuc}
\icmlauthor{Volodymyr Kindratenko}{uiuc}
\icmlauthor{ChengXiang Zhai}{uiuc}
\end{icmlauthorlist}

\icmlaffiliation{uiuc}{University of Illinois Urbana-Champaign}

\icmlcorrespondingauthor{Ke Yang}{key4@illinois.edu}
\icmlcorrespondingauthor{Volodymyr Kindratenko}{kindrtnk@illinois.edu}
\icmlcorrespondingauthor{ChengXiang Zhai}{czhai@illinois.edu}


\vskip 0.3in
]



\printAffiliationsAndNotice{}  

\input{abstract}
\input{introduction}
\input{related_work}
\input{dataset_curation_and_statistics}
\input{experiments}
\input{discussion}
\input{conclusion}
\input{limitations}
\input{ethics}

\bibliography{reference}
\bibliographystyle{icml2024}

\newpage
\appendix
\onecolumn
\input{appendix}

\end{document}

%% file: abstract.tex
\begin{abstract}
Training language models (LMs) and their application agents has become increasingly expensive due to the massive size of both datasets and models, making the cost of each test failure difficult to bear. Constructing a simplified language environment acts as a primordial training and testing ground for LMs. Such environments do not exclude commonsense knowledge or communication skill foundations, but rather present them in a plainer, more easily comprehensible form, potentially enhancing the learning efficiency of LMs, and thus reducing the required model size and data volume for effective training and evaluation. In these simplified language environments, workable strategies for small models, datasets, and agents may be adaptable to larger models, datasets, and agents in complex language environments. 

To create such a simple language environment, we propose to consider the following aspects: \textit{i)} minimizing language dataset noise and complexity, and \textit{ii)} preserving the essential characteristics of the text distribution. Prior methods have struggled with either one or both of these aspects. In contrast, we introduce a pipeline that revises text data to eliminate noise, reduce vocabulary size, and maintain genre-specific linguistic patterns (e.g., for books, conversation, code, etc.). Implementing this pipeline with large LMs, we have created a leaner suite of LM training and evaluation datasets: a 71-million token (71M) \leaner-Pretrain, a 7M \leaner-Instruct for instruction-tuning, a \leaner-\glue for assessing LMs' basic linguistic proficiency, and a \leaner-Eval for evaluating the instruction-following capability. 

Our experiments demonstrate that the leaner pre-training enhances the learning efficiency of LMs. Furthermore, we show that tiny LMs trained on the leaner datasets outperform those trained on the original dataset in following instructions across various levels of language granularity (e.g., word, sentence, etc.). Finally, since the composition of the \leaner-Pretrain aligns with the conventional one used to train large LMs, it facilitates our resource-optimized analyses of how LMs' learning objectives (e.g., masked/causal language modeling), model architectures (e.g., transformers and state space models), and training techniques (e.g., curriculum learning) affect performance on language modeling and downstream tasks. Our code and datasets are released at \href{https://github.com/EmpathYang/TinyHelen.git}{https://github.com/EmpathYang/TinyHelen.git}.

\end{abstract}

%% file: introduction.tex
\section{Introduction}

\input{Figures/ori_leaner_comparison}
\textit{``Yet at the fountainhead of rivers, the waters flow ever so slender, limpid beyond compare."}

\hfill \textit{--  Hippolyte Taine, The Philosophy of Art}

In documenting civilization, human language systems have evolved into unprecedented complexity, seemingly overlooking how early humans used simple vocabularies to build comprehensive knowledge systems. This complexity presents challenges in machine learning, especially in training large language models (LLMs) and the downstream agents in complex linguistic environments. Such training requires vast datasets and computational resources that are often unaffordable for most institutions. Additionally, many research questions, particularly those concerning model architectures and training processes, demand extensive experimentation with LLMs. The high costs of these experiments and the prohibitive loss of failures hinder progress in these critical areas.
Unlike the current methods for training LLMs, humans learn languages more efficiently by first being exposed to simplified language environments that gradually increase in complexity. This allows children, with their relatively small neural networks, to effectively grasp languages. This prompts us to explore whether we can apply this approach to LLMs by beginning with a smaller model and training it using simpler language data, thus emulating a child's language acquisition process.

It is reasonable to hypothesize that a simplified language environment, comprising of basic vocabularies and straightforward sentence structures, can efficiently train and evaluate elementary language models (LMs) and the text-based agents. It preserves core linguistic patterns while reducing complexity, enabling the use of smaller models and datasets for effective developments, which lowers the costs and thus facilitating the study of previously expensive research questions. Optimal strategies identified here (e.g., optimal model architecture or optimal training process) can potentially be applied to more complex linguistic scenarios with larger models and datasets. The tiny LM trained this way may also power text-based agents that self-evolve by following instructions and seeking information, as these skills only need basic linguistic comprehension. 

Motivated by these hypotheses, this study aims to train and evaluate tiny LMs using a novel text dataset with systematically simplified vocabularies and linguistic structures, mimicking how children learn language through simplified environments as part of their initial curriculum. Inspired by Helen Keller, we name this LM, or more precisely, the agent based on it, TinyHelen. Our work paves way for developing increasingly complex curricula, ultimately allowing language models to be trained more efficiently, akin to human language learning.

The key challenge we face is how to design the first curriculum for training TinyHelen. Given the limited capacity of tiny language models (LMs), enhancing their learning efficiency requires that the language patterns in training data are high quality and minimally essential, characterized by low noise and reduced complexity. Meanwhile, incorporating diverse pre-training dataset sources can broaden the LMs' learning scope, benefiting their generalizability \citep{longpre-etal-2024-pretrainers}. However, previous efforts to provide datasets for tiny LMs have often been limited to a single theme, such as children's stories, constraining the model's exposure to more diverse styles and knowledge \citep{eldan2023tinystoriessmalllanguagemodels,feng2024childdirectedspeecheffectivetraining}. Other datasets have not taken into account that the learning data efficiently processed by computers differs from that processed by humans, including colloquial expressions, textual noise from internet sources, or truncated expressions that lack coherence \citep{choshen2024callpapers2ndbabylm,muckatira2024emergentabilitiesreducedscalegenerative}.

In contrast, we propose a ``no noise, low complexity" principle, and apply it to design an algorithm that refines the traditional LM training data into a leaner dataset for training TinyHelen.
As illustrated by the examples from the original dataset in Figure \ref{fig:ori_leaner_comparison}, a significant portion of them, particularly web pages, contain many unwanted elements such as scrambled text, links, and advertisements. Other sources, like Wikipedia and textbooks, often feature overly detailed explanations and highly specialized facts, which obscure key linguistic patterns and basic concepts. Our revision strategy, therefore, focuses on \textit{i)} retaining essential syntactic and semantic features, \textit{ii)} eliminating irrelevant information, and \textit{iii)} simplifying the world description by rephrasing the original content using a more compact vocabulary and expressing complex ideas in simpler terms, thus reducing the information entropy of the text. In Appendix \ref{app:dataset_complexity_and_its_distribution_properties}, we demonstrate that the lower bound of the complexity of the language dataset is determined by the information entropy of its text distribution.
The language simplification of the original datasets results in \textit{i)} a 71-million token (71M) \leaner-Pretrain dataset comprising a natural language split and a formal language split, \textit{ii)} a 7M \leaner-Instruct for instruction tuning, \textit{iii)} a simplified version of the \glue benchmark \citep{wang2019gluemultitaskbenchmarkanalysis} with unchanged test topics that evaluate an LM's general linguistic capability, referred to as \leaner-\glue, and \textit{iv)} a general question-answering benchmark \leaner-Eval, consisting of 1,594 questions designed to assess the instruction-following capability of conversational models with constrained vocabulary and knowledge scope. The \leaner dataset samples are shown in Figure \ref{fig:ori_leaner_comparison} and Appendix \ref{app:additional_leaner_training_samples}. 

Our experiments intend to answer the following questions:

\textbf{How would pre-training with simplified language be beneficial?} We show that leaner pre-training with a low-complexity corpus enhances the learning efficiency of LMs. Specifically, despite its smaller size, models pre-trained with the \leaner-Pretrain outperform those trained on the original, larger corpus in downstream tasks.

\textbf{Will the simplified language help instruction-following?} We investigate whether datasets with lower noise and complexity can expedite instruction-following acquisition in tiny LMs. In particular, experiments show that models pre-trained and instruction-tuned with the leaner datasets demonstrate improvements in grammar (token-level proficiency), coherence (sentence-level proficiency), and specificity (question-answer word set relevance) when following instructions, compared to control models.

\textbf{Some questions critical to study but are constrained by cost: How do model architectures compare? What are promising curriculum learning strategies for LM pre-taining?} Given the compositional similarity between \leaner-Pretrain and conventional pre-training data, we use it for developing tiny proxy models, allowing exploring the impact of learning objectives, model architectures, and training techniques on language modeling, with reduced time and computational costs. In this phase, we control the tiny model size and pre-training data to compare the performance of LM architectures like \bert \citep{devlin2019bertpretrainingdeepbidirectional}, \llama \citep{touvron2023llamaopenefficientfoundation}, \xlnet \citep{yang2020xlnetgeneralizedautoregressivepretraining}, and \mamba \citep{gu2024mambalineartimesequencemodeling} on downstream \leaner-\glue tasks. Additionally, we explore the curriculum learning strategies for pre-training, and find that iteratively updating the training instances by LM's perplexity can save training steps and required data.

Our contributions are summarized as follows:
\vspace{-1em}
\begin{itemize}
    \item We introduce a ``no noise, low complexity" principle, supporting our proposed text revision pipeline that transforms the original training data of LMs into a version with reduced noise and complexity, which have been proven to enhance the learning efficiency of LMs.\vspace{-0.5em}
    \item We implement this pipeline to create a leaner dataset suite, the first to retain the traditional LM training dataset composition and evaluation benchmark intentions while being significantly linguistically simpler. It consists of a 71M dataset for pre-training, a 7M dataset for instruction-tuning, a benchmark that evaluates general linguistic proficiency, and a benchmark for measuring instruction-following ability. \vspace{-0.5em}
    \item Despite their infancy and limited initial performance, our instruction-following models mark one of the initial steps toward developing purely text-based self-evolving agents, an initiative we name TinyHelen. \vspace{-0.5em}
    \item The \leaner datasets offer a testing ground for low-resource scenarios, enabling us to preliminarily evaluate how model architectures and curriculum learning strategies at the pre-training stage affect language modeling performance. Experiments show that transformer decoder \llama preforms better on \leaner-\glue than state space model decoder \mamba, controlling the model size (14M parameters) and pre-training data (100M tokens). Moreover, we find iteratively introducing training instances based on the LM's perplexity can reduce both pre-training steps and data requirements.
\end{itemize}

%% file: Figures/ori_leaner_comparison.tex
\begin{figure}[t]
    \centering
    \includegraphics[width=0.5\textwidth]{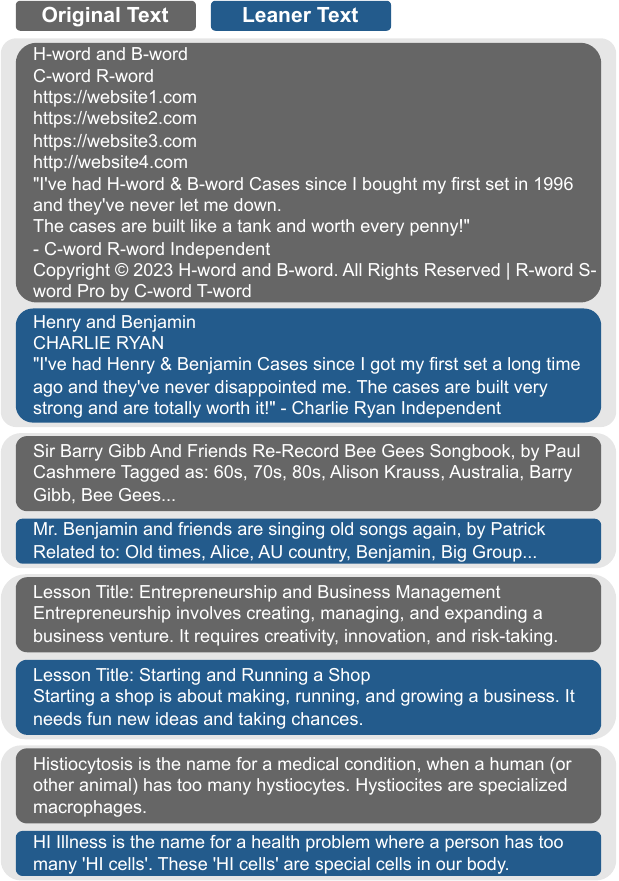}
    \vspace{-2em}
    \caption{\label{fig:ori_leaner_comparison}
        Twin samples of the original and the \leaner dataset.
    }
    \vspace{-1em}

\end{figure}

%% file: related_work.tex
\section{Related Work}
\subsection{Text-based Self-evolving Agents with Curriculum Learning}
The existing framework for developing self-evolving agents typically involves four iterative stages: experience acquisition, experience refinement, updating, and evaluation \citep{tao2024surveyselfevolutionlargelanguage}. In each cycle, agents engage with tasks to gain experience, refine it (e.g., by filtering out unsuccessful trajectories), and use the refined data for subsequent training, as showcased in work such as \citet{huang2022largelanguagemodelsselfimprove}, \citet{gulcehre2023reinforcedselftrainingrestlanguage}, \citet{wang2023selfinstructaligninglanguagemodels}, and \citet{tu2024conversationaldiagnosticai}.

However, this bootstrap approach misses the opportunity for agents to seek external insights through knowledge reflection and information requests. These agents can enhance their existing skills but struggle to improve on tasks they haven't previously encountered or successfully fulfilled.

As a complement to existing work, we propose self-evolving agents that actively seek knowledge for continual improvements, starting with tiny LMs for resource-efficient research. We introduce a no-noise, low-complexity training dataset inspired by curriculum learning \citep{soviany2022curriculumlearningsurvey}, initializing LMs' training with easier linguistic patterns to smooth the loss function and improve their learning efficiency.

\subsection{Tiny Datasets for Tiny Language Models}
\input{Tables/tiny_datasets}
Recent studies have focused on developing or compiling tiny-scale datasets to explore language modeling within a dataset size similar to children's mind development and support research in low-resource contexts. These studies also train tiny-scale models to validate their datasets, detailed in Table \ref{table:tiny_datasets}. Specifically, \citet{eldan2023tinystoriessmalllanguagemodels} create a dataset of children's stories using only words understandable by children, alongside models capable of generating coherent and logical children's stories. Additionally, \citet{feng2024childdirectedspeecheffectivetraining} introduces an LLM-synthesized children's dialogue dataset to examine the impact of age and discourse ordering in the dataset on training efficiency. Furthermore, \citet{warstadt2023papersbabylmchallenge} and \citet{choshen2024callpapers2ndbabylm} have compiled a dataset focusing on subtitles, child-directed speech, and simple Wikipedia articles for the BabyLM challenge, aiming to explore tiny-scale pre-training and cognitive modeling. Lastly, \citet{muckatira2024emergentabilitiesreducedscalegenerative} explores how data simplification through filtering web data truncations using a vocabulary of 21k+ words can induce emergent abilities in tiny models. 

However, none of these studies have resolved optimizing LM training data to reduce noise and complexity while retaining the diversity of data sources. Such limitations are addressed in our new dataset.

\subsection{Model Architecture Comparison and Curriculum Learning Strategies for Pre-training}
Several studies compare transformers and state space models. \citet{patro2024mamba360surveystatespace} examines transformer and state space model variants on computer vision (CV) and natural language processing (NLP) benchmarks but doesn't control for model size or pre-training data. \citet{waleffe2024empiricalstudymambabasedlanguage} controls for these variables (8B parameters, 3.5T tokens) when comparing transformers and \mamba models on tasks like commonsense reasoning and long-text tasks, finding \mamba models can match or exceed transformers. In contrast, our study uses a simpler language environment with smaller models (14M parameters) and a smaller dataset (100M tokens), focusing on evaluating general linguistic ability, which leads to different conclusions.

Regarding curriculum learning for causal language modeling pre-training, \citet{campos2021curriculumlearninglanguagemodeling} uses sample-difficulty-measuring metrics like sentence length and entropy, applying competence-based sampling but finds no significant training improvements. \citet{li2022stabilityefficiencydilemmainvestigatingsequence} and \citet{nagatsuka2023length} use sentence length, concluding that curriculum learning accelerates convergence, reducing token use and training time with a step-wise linear pacing function and the Square scheduler, respectively. \citet{fan2023irreduciblecurriculumlanguagemodel} employs a learnability score, i.e., the loss gap from a proxy model at early and late stages, prioritizing samples with high scores, achieving better validation perplexity and accuracy on knowledge benchmarks. Our study, conducted in a simpler linguistic setting, uses metrics like sentence length, LM self-paced loss, and proxy LM loss. Our pacing function iteratively updates the training dataset to be sampled. We evaluate performance on general linguistic tasks.

%% file: Tables/tiny_datasets.tex
\begin{table}[ht]
\centering
\caption{Tiny datasets and their trained tiny models.}
\label{table:tiny_datasets}
\setlength{\tabcolsep}{8pt}
\renewcommand{\arraystretch}{1.2}
{\scriptsize
\begin{tabular}{
lp{0.098\textwidth}p{0.092\textwidth}
}
\toprule
\textbf{Work}     & \textbf{Dataset Size (M)} & \textbf{Model Size (M)} \\ \midrule
TinyStories \citep{eldan2023tinystoriessmalllanguagemodels}         & 472                      & 1-33 \\
TinyDialogues \cite{feng2024childdirectedspeecheffectivetraining}       & 29                       & 124, 125                \\
BabyLM \citep{choshen2024callpapers2ndbabylm}              & 100                      & 10, 100                 \\
miniGPT \citep{muckatira2024emergentabilitiesreducedscalegenerative}             & 2100, 22000              & 1-165                   \\ \midrule
TinyHelen               & 71                       & 1, 14                   \\
\bottomrule
\end{tabular}}
\end{table}

%% file: dataset_curation_and_statistics.tex
\section{Dataset Curation and Statistics}
Our goal is to refine the original (naturally available) training data of large LMs, such as web text, code, and instruction-answer pairs, by reducing noise and simplifying the text for easier learning. As demonstrated in Appendix \ref{app:dataset_complexity_and_its_distribution_properties}, the complexity of a language dataset is lower-bounded by the information entropy of its text distribution. Thus, our dataset revision process focuses on reducing vocabulary size and rephrasing content in simpler terms to minimize entropy. The entire pipeline is outlined in Algorithm \ref{alg:leaner_training}. Additionally, we revise the \glue \citep{wang2019gluemultitaskbenchmarkanalysis} benchmark for testing LMs' linguistic abilities, and create an instructing-following benchmark adhering to the same principle, presented in Algorithm \ref{alg:leaner_glue} and \ref{alg:leaner_eval}, respectively. The key algorithm components are elaborated on in the following sections. We name the initial datasets \leaner-Pretrain-71M, \leaner-Instruct-7M, \leaner-\glue, and \leaner-Eval-1K, omitting the token counts for brevity in later references.

\subsection{\leaner-Pretrain and \leaner-Instruct}
\input{Algorithms/leaner_training}
\subsubsection{Data Sources}
We follow a pretrainer's guide \citep{longpre-etal-2024-pretrainers} when deciding the pre-training corpus for simplification, including standard domains like web, books, wikis, and conversations for natural language, and code and math for formal language. We exclude specialized areas like biomed and academic texts due to their atypical distribution. Given the proven benefits of high-quality textbooks \citep{gunasekar2023textbooksneed}, we also include a textbook segment in the natural language portion. We randomly sample instances from the following sources for each data category until reaching the maximum token number specified in Algorithm \ref{alg:leaner_training}. For natural language, \textsc{cerebras/SlimPajama-627B} \citep{cerebras2023slimpajama} provides web and book data, \textsc{rahular/simple-wikipedia} \citep{simplewiki} is used for the wiki subset, \textsc{nampdn-ai/tiny-textbooks} \citep{nam_pham_2023} for textbooks, and \textsc{opensubtitles} \citep{lison-tiedemann-2016-opensubtitles2016} for conversations. For formal language, \textsc{nampdn-ai/tiny-codes} \citep{tinycodes} is used for the code subset, and \textsc{MATH amps} \citep{hendrycksmath2021} for math. We rank and partition the formal language subsets by difficulty before sampling from them, as detailed in Appendix \ref{app:formal_language_difficulty_ranking}. We only begin sampling from a more difficult group after exhausting the easier ones. Last, we refine the instruction-tuning dataset from \textsc{yahma/alpaca-cleaned} \citep{yahmaalpacacleaned}. The statistics of the original datasets are shown in Table \ref{table:original_training_statistics}.

\subsubsection{Language Simplification}
\input{Tables/leaner_training_statistics}
\input{Tables/ngram_entropy}
To manage time and cost constraints, we rely on LLMs rather than human annotators to revise datasets. Specifically, the LLM is prompted to rephrase text, ensuring simple grammar and a vocabulary limited to the 2,000 most common words, and to simplify complex concepts. Furthermore, for noisy sources like web data, we instruct the LLMs to exclude irrelevant content such as formatting and advertisements. Detailed prompts are in Appendix \ref{app:prompts_for_creating_leaner_training_leaner_glue_and_leaner_eval}, with the heuristics behind provided in the later paragraphs. 

The main challenge in simplifying natural language and instruction splits lies in overly detailed descriptions, which often include numerous references to specific entities such as individuals, places, organizations, times, artifacts, and links. Retaining these references would overextend the vocabulary and overwhelm a tiny model with trivial and detailed relationships, illustrated by samples such as ``The Alexander ALX400 (later known as the TransBus ALX400 and the Alexander Dennis ALX400) is a 2-axle double-decker low-floor bus body. It was built from 1997 to 2006 by Alexander and then TransBus International and then Alexander Dennis respectively.'' Therefore, \textit{we simplify the world setting}, assuming a limited number of people from a small set of places or organizations, at general times, engaged in simple activities. The guidelines to simplify the world setting are summarized below.

\begin{tcolorbox}[colback=leaner!5!white,colframe=leaner!80!black,title=World Setting Simplification Guidelines,fonttitle=\small]
{\small
i. Replace any mentioned name with a name from the provided list that shares the same initial letter.\\
ii. To rephrase an entity like a place, use its first two uppercase letters followed by its name.\\
iii. Only keep specific times if essential for understanding the context; otherwise, use vague references.\\
iv. Use simple, common language to describe an artifact.}
\end{tcolorbox}

Moreover, for formal language splits (code and math), ranking and sampling by difficulty ensures simplicity, reducing the need for extensive alterations. We only rephrase task descriptions, comments, and explanations in the code subset, keeping math problems and solutions unchanged.

We input the prompt augmented with the text to be revised, collect the LLM's outputs, and compile them into a dataset. If a text exceeds the LLM's max sequence length, we divide it, revise each part independently, and then concatenate the results. We employ \textsc{gpt-4-0125} \cite{achiam2023gpt} for this revision process. Table \ref{table:leaner_training_statistics} presents the leaner dataset statistics, while Figure \ref{fig:ori_leaner_comparison} and Appendix \ref{app:additional_leaner_training_samples} show examples. The n-gram entropy for both the original and the \leaner datasets is computed and presented in Table \ref{table:ngram_entropy}, which indicates that the \leaner datasets have a lower entropy compared to the original datasets.
\subsection{\leaner-\glue}
\label{subsec:leaner_glue}
\input{Algorithms/leaner_glue}
We intend to evaluate the performance of tiny models across a diverse range of tasks, with the \glue benchmark \citep{wang2019gluemultitaskbenchmarkanalysis} serving as a promising candidate. \glue is a suite of tasks designed to assess a model’s general linguistic capabilities, covering areas such as linguistic acceptability (CoLA \citep{warstadt-etal-2019-neural}), sentiment analysis (SST2 \citep{socher-etal-2013-recursive}), sentence similarity (MRPC \citep{dolan-brockett-2005-automatically}, QQP \citep{qqponlinedataset}, and STS-B \citep{cer-etal-2017-semeval}), and natural language inference (MNLI \citep{williams-etal-2018-broad}, QNLI \citep{rajpurkar2016squad100000questionsmachine,wang2019gluemultitaskbenchmarkanalysis}, RTE \citep{Dagan2005ThePR,BarHaim2006TheSP,giampiccolo-etal-2007-third,DBLP:conf/tac/BentivogliMDDG09}, and WNLI \citep{Levesque2011TheWS}). Each task includes training, validation, and test sets.

The datasets contain complex samples, such as ``Thebes held onto power until the 12th Dynasty, when its first king, Amenemhet I, who reigned between 1980–1951 B.C., established a capital near Memphis." To maintain a consistent vocabulary and knowledge scope across the tiny LMs' training and evaluation, thus enabling potential zero-shot evaluation of them, we have simplified the \glue benchmark. In revising the text, we adhere to principles used in rephrasing the \leaner-Pretrain and \leaner-Instruct datasets: downsizing vocabulary to reduce linguistic complexity and expressing complex concepts in simpler terms. Full prompts are in Appendix \ref{app:prompts_for_creating_leaner_training_leaner_glue_and_leaner_eval}, with statistics shown in Table \ref{table:leaner_glue_statistics}.

Recognizing that simplifications might alter the original testing intentions, we ensure consistency between task inputs and labels through multiple rounds of verification and regeneration until a significant portion of the dataset is covered. Specifically, we utilize LLMs to predict the labels of test instances and retain samples only if the model’s predictions match the true labels. We continue to revise and evaluate the unqualified samples to expand the retained set until we achieve our target coverage ratio $r=70\%$. Furthermore, since the original datasets do not provide labels for the test set, we randomly split the revised validation set into two distributionally similar validation and test sets.

We would like to emphasize that our algorithm is versatile and can potentially convert any NLP benchmark collection into a leaner version.

\subsection{\leaner-Eval}
\label{subsec:leaner_eval}
\input{Algorithms/leaner_eval}
The existing benchmarks for testing the instruction-following capabilities of LMs are primarily designed for LLMs with extensive vocabularies and knowledge bases. To address this, we propose a set of problems and evaluation metrics named \leaner-Eval, specifically for assessing the instruction-following abilities of tiny LMs with limited vocabulary and knowledge scope. 
During our dataset curation, we aim to generate questions that test the linguistic proficiency and commonsense knowledge of LMs, without delving into advanced knowledge-based queries such as the mechanism behind the hexagonal symmetry of snowflakes. To ensure a diverse and relevant set of questions, we sample examples from a question set that features a broad range of topics in simple language. These examples serve as templates for question structure and content, and are incorporated into the prompts to guide the generation of new questions.
Therefore, our problem generation strategy involves randomly sampling 10 questions from the \leaner-Instruct dataset as examples, then using LLMs (\textsc{gpt-4-0125} and \textsc{gpt-4o-mini}) to generate 50 varied and simple new questions, repeating this process 50 times (full prompt details are provided in Appendix \ref{app:prompts_for_creating_leaner_training_leaner_glue_and_leaner_eval}). From the collected questions, we use the \textsc{SimCSE} tool \citep{gao2022simcsesimplecontrastivelearning}, which measures sentence embedding similarity, to remove semantically duplicated questions. We discard questions with a cosine similarity greater than 0.876, a threshold empirically determined to be effective, resulting in a final set of 1,594 questions. Last, these questions are assessed using \textsc{SimCSE} regarding their similarity to the \leaner- and Original-Instruct datasets, and are categorized into four similarity ranges (instance counts in parentheses): 0-0.7 (214), 0.7-0.8 (604), 0.8-0.9 (535), and 0.9-1 (241), based on their maximum similarity to any question in the instructing-tuning data.

The evaluation metrics for instruction-following generations are \textit{grammar} (ensuring responses are grammatically correct, testing the language model's understanding of token relationships), \textit{coherence} (checking for logical structure and smooth flow without repetition, assessing sentence-level linguistic proficiency), and \textit{specificity} (ensuring responses are directly relevant, factually accurate, and plausible, evaluating the model's ability to to align its output with the question's content). Each metric is scored on an integer scale from 0 to 5 (where 0 is the worst and 5 is the best), making the total score range for an LM's response between 0 and 15. We use \textsc{gpt-4-0125} as the evaluator, scoring the responses from the tiny LMs, with the full evaluation prompt provided in Appendix \ref{app:prompts_for_creating_leaner_training_leaner_glue_and_leaner_eval}. We shuffle the responses before testing to mitigate potential position bias in the LLMs.

%% file: Algorithms/leaner_training.tex
\begin{algorithm}[ht]
   \caption{\leaner-Training Dataset Collection}
   \label{alg:leaner_training}
\begin{algorithmic}
   \STATE {\bfseries Input:}
   \STATE Pre-training size: 100M tokens, distributed as:
   \STATE \quad Natural language ($NL$): \{web: 65\%, book: 5\%, wiki: 2.5\%, textbook: 5\%, conversation: 2.5\%\};
   \STATE \quad Formal language ($FL$): \{code: 15\%, math: 5\%\}.
   \STATE Instruction-tuning ($Instruct$) size: 10M tokens.
   \STATE {\bfseries Procedure:}
   \STATE 1. Identify sources for each subset.
   \STATE 2. {\bfseries for} $subset$ {\bfseries in} $NL$ {\bfseries do}
   \STATE \quad \quad Randomly sample from source until target token count is met.
   \STATE \quad {\bfseries end for}
   \STATE 3. {\bfseries for} $subset$ {\bfseries in} $FL$ {\bfseries do}
   \STATE \quad \quad Randomly sample from source partitions, ordered by increasing difficulty, until target token count is met.
   \STATE \quad {\bfseries end for}
   \STATE 4. {\bfseries for} $subset$ {\bfseries in} $NL \cup FL \cup Instruct$ {\bfseries do}
   \STATE \quad \quad Define revision principles.
   \STATE \quad \quad An LLM revises text following principles.
   \STATE \quad {\bfseries end for}
\end{algorithmic}
\end{algorithm}

%% file: Tables/leaner_training_statistics.tex
\begin{table*}[t]
\centering
\caption{\leaner-Pretrain and \leaner-Instruct statistics.}
\label{table:leaner_training_statistics}
\renewcommand{\arraystretch}{1.2} 
{\scriptsize
\begin{tabular}{l|c|ccccccc|c}
\toprule
\textbf{Unit: M tokens} & \textbf{\leaner-Pretrain} & \textbf{web} & \textbf{book} & \textbf{wiki} & \textbf{textbook} & \textbf{conversation} & \textbf{code} & \textbf{math} & \textbf{\leaner-Instruct} \\ 
\midrule
\textbf{All}        & 70.63 & 42.74 & 4.04 & 2.34 & 5.26 & 1.00 & 11.27 & 3.99 & 6.86 \\
\textbf{Train}      & 64.67 & 39.38 & 3.56 & 2.12 & 4.88 & 0.90 & 10.23 & 3.60 & 6.72 \\
\textbf{Validation} & 3.00  & 1.67  & 0.25 & 0.11 & 0.19 & 0.05 & 0.53  & 0.20 & 0.07 \\
\textbf{Test}       & 2.97  & 1.69  & 0.23 & 0.11 & 0.18 & 0.05 & 0.51  & 0.20 & 0.07 \\
\bottomrule
\end{tabular}}
\end{table*}

%% file: Tables/ngram_entropy.tex
\begin{table*}[ht]
\centering
\caption{N-gram entropy of the \leaner-Training and the Original-Training datasets.}
\label{table:ngram_entropy}
\setlength{\tabcolsep}{8pt}
\renewcommand{\arraystretch}{1.2}
{\scriptsize
\begin{tabular}{c|cc|cc|cc|cc}
\toprule
\multirow{2}{*}{\textbf{N-gram}} & \multicolumn{2}{c|}{\textbf{All}}       & \multicolumn{2}{c|}{\textbf{Natural Language}} & \multicolumn{2}{c|}{\textbf{Formal Language}} & \multicolumn{2}{c}{\textbf{Instruction}} \\ 
\cline{2-9} 
                        & \textbf{Ori}   & \textbf{\leaner} & \textbf{Ori}       & \textbf{\leaner}    & \textbf{Ori}      & \textbf{\leaner}    & \textbf{Ori}    & \textbf{\leaner} \\ 
\midrule
1                       & 16.41 & 15.86 & 16.27 & 15.55 & 15.48 & 15.00 & 15.64 & 14.94 \\ 
2                       & 22.76 & 21.16 & 22.58 & 20.72 & 19.79 & 19.33 & 19.68 & 18.87 \\ 
3                       & 24.51 & 23.25 & 24.35 & 22.85 & 21.29 & 20.95 & 21.19 & 20.57 \\
\bottomrule
\end{tabular}}
\end{table*}

%% file: Algorithms/leaner_glue.tex
\begin{algorithm}[ht]
   \caption{\leaner-\glue Benchmark Creation}
   \label{alg:leaner_glue}
\begin{algorithmic}
   \STATE {\bfseries Input:}
   \STATE Minimum coverage ratio $r=70\%$.
   \STATE {\bfseries Procedure:}
   \STATE Define revision principles.
   \FOR{$task$ {\bfseries in} \glue~}
   \STATE Load the original sample list $ori$.
   \STATE Initialize the leaner sample list $leaner=[]$.
   \WHILE{len($leaner$) $<$ $r \times$len($ori$)}
   \STATE Keep a revised sample buffer $revised=[]$.
   \FOR{$sample$ {\bfseries in} $ori\backslash leaner (pre-revision)$}
   \STATE An LLM revises $sample$ following principles and outputs $revised\_sample$.
   \STATE $revised$ $+=$ $[revised\_sample]$.
   \ENDFOR
   \FOR{$sample$ {\bfseries in} $revised$}
   \STATE Given the input $x$ and label $y\_true$ in $sample$, let an LLM predict $x$'s label $y\_pred$.
   \IF{$y\_pred$ $==$ $y\_true$}
   \STATE $leaner$ $+=$ $[sample]$.
   \ENDIF
   \ENDFOR
   \ENDWHILE
   \STATE {\bfseries document} $leaner$.
   \ENDFOR
\end{algorithmic}
\end{algorithm}

%% file: Algorithms/leaner_eval.tex
\begin{algorithm}[ht]
   \caption{\leaner-Eval Benchmark Creation}
   \label{alg:leaner_eval}
\begin{algorithmic}
   \STATE {\bfseries Input:}
   \STATE Generation iteration $I=50$, example number $n_e=10$, output question number $n_q=50$, similarity threshold $s=0.876$.
   \STATE {\bfseries Procedure:}
   \STATE Initialize the raw leaner question list $raw\_Qs=[]$.
   \FOR{$i$ {\bfseries in} $I$}
   \STATE Instantiate the prompt with $n_e$ randomly sampled examples from \leaner-Instruct, and specify $n_q$.
   \STATE An LLM generates the question list $qs$.
   \STATE $raw\_Qs$ $+=$ $qs$.
   \ENDFOR
   \STATE Deduplicate $raw\_Qs$ to $Qs$ by question similarity $>s$.
   \FOR{$q$ {\bfseries in}  $Qs$}
   \STATE Compute $q$'s similarity with instances in \leaner- and Ori-Instruct, and record them.
   \ENDFOR
   \STATE {\bfseries document} $Qs$.
\end{algorithmic}
\end{algorithm}

%% file: experiments.tex
\section{Experiments}

The main goal of our experiments is to study the following three new research questions around the optimization of efficiency and effectiveness of LLM training:   
\textit{i)} Can training with clean datasets that have lower linguistic complexity (e.g., \leaner-Training) enhance the learning efficiency of LMs? \textit{ii)} Do LMs pre-trained and instruction-tuned with these low-complexity datasets tend to develop instruction-following abilities earlier? and \textit{iii)} Would the \leaner datasets, similar in composition to traditional LLM training sets, enable more efficient development of LM architectures and training techniques on a resource-efficient scale?

To answer these questions, we design three experiments. Experiment 1 (\explm) compares downsized versions of varied LM architectures on language modeling and downstream tasks, ensuring consistent pre-training data and model size. Experiment 2 (\expinstruct) assesses the instruction-following abilities of tiny LMs trained with different pre-training and instruction-tuning data combinations. Experiment 3 (\expcurriculum) explores curriculum learning strategies for LM pre-training.
\input{exp-lm}
\input{exp-instruct}
\input{exp-curriculum}

%% file: exp-lm.tex
\subsection{\explm: Comparing Model Architectures with the \leaner Dataset Suite}

\subsubsection{Models}
\label{subsubsec:models}
In this experiment, we control for model size, training data, and setup to compare the learning efficiency of different LM architectures. Specifically, we select LM architectures including \bert (transformer \citep{vaswani2023attentionneed} encoder) \citep{devlin2019bertpretrainingdeepbidirectional}, \llama (transformer decoder) \citep{touvron2023llamaopenefficientfoundation}, \xlnet (which uses permutation-based language modeling) \citep{yang2020xlnetgeneralizedautoregressivepretraining}, and \mamba (state space model) \citep{gu2024mambalineartimesequencemodeling}.\footnote{We exclude \textsc{T5} (full transformer) \citep{raffel2023exploringlimitstransferlearning} because, to ensure a fair comparison, we reset the tokenizer vocabulary size of all models to 2K tokens. \textsc{T5}'s special tokens occupy 5\% of the total vocabulary, an order of magnitude more than in other models, disabling control over the effective vocabulary.} We configure each model to have approximately 14M parameters, with detailed parameters provided in Appendix \ref{app:explm_settings} and model sizes listed in Table \ref{table:exp1-leaner_glue}. To ensure comparable tokenization, we limit the vocabulary size to 2K tokens and retrain new tokenizers using the \leaner-Training corpus.  All tokenizers are initialized with the \textsc{OpenAI-community/GPT-2} \cite{radford2019language} tokenizer. For \llama and \mamba, both decoders, the retrained tokenizer could be applied directly. For \bert and \xlnet, we replace less common words in the retrained tokenizer with model-specific special tokens (5 for \bert and 9 for \xlnet).

The pre-training datasets included the natural language sections of \leaner-Pretrain and Original-Pretrain. \bert, \llama, and \xlnet are trained using model code from HuggingFace \citep{wolf-etal-2020-transformers}, while Mamba is trained using the GitHub \textsc{state-spaces/mamba} \citep{mamba2024repo}. Detailed machine and training settings are in Appendix \ref{app:explm_settings}. We perform a grid search of leanrning rate to identify the setup with the lowest evaluation loss for each architecture. The model hyperparameter selection is also provided in Appendix \ref{app:explm_settings}.

\subsubsection{Benchmark}
\label{subsubsec:explm_benchmark}
Beyond language modeling, we assess the performance of these pre-trained tiny LMs against their non-pretrained counterparts on general linguistic downstream tasks using the \leaner-\glue benchmark (introduced in Section \ref{subsec:leaner_glue}). \leaner-\glue is a simplified vocabulary version of the \glue benchmark \citep{wang2019gluemultitaskbenchmarkanalysis}, maintaining the original test objectives, which include:

\textbf{Linguistic acceptability}: \leaner-CoLA; \\
\textbf{Sentiment analysis}: \leaner-SST2; \\
\textbf{Sentence similarity}: \leaner-MRPC, \leaner-QQP, and \leaner-STS-B; \\
\textbf{Natural language inference}: \leaner-MNLI, \leaner-QNLI, \leaner-RTE, and \leaner-WNLI.

The training, validation, and test dataset sizes for this benchmark are shown in Table \ref{table:leaner_glue_statistics}. We perform a grid search to determine the optimal hyperparameter settings (e.g., learning rate) for each model on each subtask of the benchmark. Each experiment under a specific hyperparameter configuration is run five times with different random seeds. We calculate the test performance of the model with the best average validation performance as the model's score for each benchmark task. Furthermore, we min-max normalize the model's scores across subtasks and compute the mean of all normalized performance scores to represent the model's average performance.

\subsubsection{Results}
\label{subsubsec:explm_results}
\input{Tables/exp1-leaner_glue}
The results are presented in Table \ref{table:exp1-leaner_glue}. Firstly, it is notable that providing a simpler language environment as the learning context enhances the learning efficiency of LMs. This is evidenced by the fact that all LMs pre-trained with the \leaner-Pretrain dataset outperform those pre-trained with the Original-Pretrain dataset in terms of average scores and most sub-task scores, despite \leaner-Pretrain is 41\% smaller (71M tokens compared to 100M tokens). Secondly, pre-training offers LMs a significant advantage on downstream tasks, as pre-trained models consistently outperform those fine-tuned from scratch on each sub-task. Finally, the learning efficiency varies among different LM architectures. Under both pre-training scenarios, the order is \xlnet $>$ \bert $>$ \llama $>$ \mamba, while for models fine-tuned on sub-task from scratch, the order is \llama $>$ \xlnet $>$ \mamba $>$ \bert. It is important to note that \underline{\textbf{LM Loss}} is not comparable across LM architectures, except \llama and \mamba, due to differing learning objectives. Additionally, because the epoch-wise learning objective remains constant for decoders (i.e., \llama and \mamba), they are more prone to overfitting, resulting in fewer pre-training steps for the final tested models compared to other architectures.

%% file: Tables/exp1-leaner_glue.tex
\begin{table*}[ht]
\centering
\caption{\leaner models' performance on the \leaner-\glue benchmark. \underline{\textbf{Size (M)}} shows the model size w/o and w/ learnable word embeddings. \underline{\textbf{LM Loss}} reports the minimum training loss. All \glue results are averaged over 5 random seeds, tested with the best models selected via grid search over hyperparameters. \underline{\textbf{Avg}} displays the average of the column-wise min-max normalized \glue scores.}
\label{table:exp1-leaner_glue}
\setlength{\tabcolsep}{3pt}
\renewcommand{\arraystretch}{1.2}
{\scriptsize
\begin{tabular}{l|l|c|c|cccccccc|c}
\toprule
\multirow{2}{*}{\textbf{Initialization}} & \multirow{2}{*}{\textbf{Model}} & \multirow{2}{*}{\textbf{Size (M)}} & \multirow{2}{*}{\textbf{LM Loss}} & \multicolumn{9}{c}{\textbf{\leaner-\glue\footnotemark}} \\ \cline{5-13} 
                                 &                         &                           &                          & \textbf{CoLA} & \textbf{SST2} & \textbf{MRPC} & \textbf{QQP} & \textbf{STS-B} & \textbf{MNLI} & \textbf{QNLI} & \textbf{RTE} & \textbf{Avg} \\ \midrule
\multirow{4}{*}{\textbf{\leaner Pre-trained}} & \bert & 13.68 / 14.79 & 0.9 & \best{38.2} & 88.4 & 77.6 & 86.6 & 76.5 & 77.3 & \best{84.4} & 57.3 & 0.861 \\
                                   & \llama & 13.79 / 14.43 & 2.3 & 24.5 & 88.7 & 72.6 & 84.4 & 67.0 & 75.3 & 80.7 & 53.6 & 0.678 \\
                                   & \xlnet & 14.06 / 14.78 & 2.9 & 34.5 & \better{89.2} & \best{85.3} & \best{87.5} & \best{80.6} & \best{79.8} & \best{84.4} & \better{60.0} & \best{0.963} \\
                                   & \mamba & 13.93 / 15.40 & 2.3 & 25.2 & 88.3 & 68.9 & 84.8 & 59.8 & 73.3 & 75.5 & 52.2 & 0.590 \\ \midrule
\multirow{4}{*}{\textbf{Randomly Initialized}} & \bert & 13.68 / 14.79 & - & \better{14.1} & 76.8 & 68.8 & 71.2 & 18.1 & 60.7 & 61.9 & 48.5 & 0.071 \\
                                   & \llama & 13.79 / 14.43 & - & 12.9 & 80.9 & 69.2 & \better{80.7} & \better{35.5} & 66.1 & \better{67.1} & 51.8 & \better{0.301} \\
                                   & \xlnet & 14.06 / 14.78 & - & 11.0 & 77.2 & \better{72.6} & 77.1 & 17.4 & \better{68.3} & 60.7 & \better{52.8} & 0.212 \\
                                   & \mamba & 13.93 / 15.40 & - & 7.1 & \better{84.8} & 69.6 & 74.6 & 14.0 & 61.3 & 58.9 & 46.3 & 0.118 \\ \midrule
\multirow{4}{*}{\textbf{Ori Pre-trained}}  & \bert & 13.68 / 14.79 & 1.1 & \better{31.0} & 85.9 & 76.8 & 86.0 & 74.0 & 75.0 & 82.7 & 52.1 & 0.729 \\
                                   & \llama & 13.79 / 14.43 & 2.7 & 20.1 & 83.1 & 72.7 & 83.2 & 65.6 & 72.3 & 79.5 & 52.2 & 0.558 \\
                                   & \xlnet & 14.06 / 14.78 & 3.1 & 22.3 & \best{89.5} & \better{81.7} & \better{87.0} & \better{80.0} & \better{79.4} & \better{83.9} & \best{62.5} & \better{0.900} \\
                                   & \mamba & 13.93 / 15.40 & 2.5 & 22.6 & 88.3 & 68.3 & 83.0 & 32.1 & 71.7 & 68.7 & 52.8 & 0.470 \\ \bottomrule
\end{tabular}}
\end{table*}

\afterpage{\footnotetext{Due to the limited number of training and evaluation samples for the \leaner WNLI, totaling only 31, there is a substantial variance that hinders drawing significant average conclusions. Thus, we have chosen not to report the tiny LMs' performance on WNLI.}}

%% file: exp-instruct.tex
\subsection{\expinstruct: Evaluating the Instruction-following Ability of Tiny LMs Trained with Different Data Recipes}

\subsubsection{Models and the Benchmark}
We employ two pre-training datasets and two instruction-tuning datasets, each in both leaner and original versions, creating four data recipes for training instruction-following models. We train the \llama model with 14M parameters using consistent settings, with details on model configuration, hardware, training settings, and hyperparameters provided in Appendix \ref{app:expinstruct_settings}. During pre-training, we select the model with the lowest evaluation loss from both the \leaner-Pretrain and Original-Pretrain datasets as the starting point for instruction-tuning. In the instruction-tuning phase, we train the models to convergence with the same number of steps. Since the instruction-tuning evaluation loss only indicates the perplexity with reference answers but not instruction-following ability, we randomly sample 31 benchmark questions for preliminary evaluation. The best-performing model for each data recipe is chosen to represent that group of models. This results in four model variants named according to the training dataset recipes: \leaner-\leaner, \leaner-Ori, Ori-\leaner, and Ori-Ori. 

To compare those tiny LMs, we use \leaner-Eval (introduced in Section \ref{subsec:leaner_eval}), a benchmark consists of 1,594 questions assessing the instruction-following ability of conversational models with limited vocabulary and knowledge. Evaluation metrics include grammar, coherence, and answer specificity. Additionally, the benchmark is partitioned by question similarity to instances in the instruction-tuning datasets, enabling generalizability assessment.

\subsubsection{Results}
\label{subsubsec:expinstruct_results}
\input{Tables/exp2-instruction_following_evaluation}
The evaluation results of the model's responses are listed in Table \ref{table:exp2-instruction_following_evaluation}, with examples of instruction-following generation provided in Appendix \ref{app:instruction_following_examples}. First, we observe that a 14M LM pre-trained on approximately 70M tokens struggles to consistently produce high-quality responses. This can be inferred from the optimal model's average metric-wise score of around 1, indicating a \textit{bad} level according to the benchmark. Notably, the 100M token dataset is comparable to the total volume of text accessible to a human adolescent \citep{Gilkerson2017MappingTE}, which guides our training dataset size decision. However, we acknowledge that during early human development, individuals are exposed to multimodal data, and the functioning of the human brain differs from computational models. Consequently, a 100M token scale \leaner-Pretrain dataset is insufficient for enabling a 10M scale model (larger models would easily overfit on 100M tokens) to acquire instruction-following capabilities.

Furthermore, when analyzing the overall performance by holding constant the instruction-tuning corpus, we find that pre-training on the \leaner dataset enhances the grammar, coherence, and specificity of instruction-following generations, with grammar improvements being the most pronounced. Additionally, when comparing model performance across different similarity partitions, we notice a significant decline in response quality for questions without similar instances in the instruction-tuning training set.

Finally, the \leaner-Ori model outperforms the \leaner-\leaner model. We attribute it to the \leaner-ori model's tendency to produce general responses, such as ``As an AI, I do not have the capability to generate or organize information...", when faced with uncertainty, unlike other models that sometimes respond with incoherent token sequences.

%% file: Tables/exp2-instruction_following_evaluation.tex
\begin{table*}[ht]
\centering
\caption{Evaluation of 14M \textsc{Llamas}' instruction-following ability trained with different data recipes.}
\label{table:exp2-instruction_following_evaluation}
\setlength{\tabcolsep}{3pt}
\renewcommand{\arraystretch}{1.2}
{\scriptsize
\begin{tabular}{c|cccc|cccc}
\toprule
\multirow{2}{*}{\begin{tabular}[c]{@{}c@{}}Model's Pre-training \\ \& Instruction-tuning \\ Data Recipe\end{tabular}} & \multicolumn{4}{c|}{\textbf{All}} & \multicolumn{4}{c}{\textbf{Partitions by Training Similarity (Metric: Sum)}} \\ \cline{2-9} 
                                   & \begin{tabular}[c]{@{}c@{}}\textbf{Grammar}\\ (/5)\end{tabular} & \begin{tabular}[c]{@{}c@{}}\textbf{Coherence}\\ (/5)\end{tabular} & \multicolumn{1}{c|}{\begin{tabular}[c]{@{}c@{}}\textbf{Specificity}\\ (/5)\end{tabular}} & \begin{tabular}[c]{@{}c@{}}\textbf{Sum}\\ (/15; \#1594)\end{tabular} & \begin{tabular}[c]{@{}c@{}}\textbf{0-0.7}\\ (/15; \#214)\end{tabular} & \begin{tabular}[c]{@{}c@{}}\textbf{0.7-0.8}\\ (/15; \#604)\end{tabular} & \begin{tabular}[c]{@{}c@{}}\textbf{0.8-0.9}\\ (/15; \#535)\end{tabular} & \begin{tabular}[c]{@{}c@{}}\textbf{0.9-1}\\ (/15; \#241)\end{tabular} \\ \midrule
\textbf{Ori}-Ori            & 1.1 & 0.9 & \multicolumn{1}{c|}{0.6} & 2.7 & 2.1 & 2.8 & 2.8 & 2.7 \\
\textbf{\leaner}-Ori         & 1.7 \improvement{0.6} & 1.4 \improvement{0.5} & \multicolumn{1}{c|}{1.0 \improvement{0.4}} & 4.1 \improvement{1.4} & 3.4 \improvement{1.3} & 4.2 \improvement{1.4} & 4.1 \improvement{1.3} & 4.2 \improvement{1.5} \\ \midrule
\textbf{Ori}-\leaner         & 1.3 & 1.1 & \multicolumn{1}{c|}{0.8} & 3.1 & 2.5 & 3.2 & 3.1 & 3.4 \\
\textbf{\leaner}-\leaner      & 1.5 \improvement{0.2} & 1.2 \improvement{0.1} & \multicolumn{1}{c|}{0.9 \improvement{0.1}} & 3.5 \improvement{0.4} & 3.3 \improvement{0.8} & 3.6 \improvement{0.4} & 3.6 \improvement{0.5} & 3.8 \improvement{0.4} \\
\bottomrule
\end{tabular}
}
\end{table*}

%% file: exp-curriculum.tex
\subsection{\expcurriculum: Investigate Curriculum Learning Strategies for Pre-training with Tiny Proxy Models}
\input{Figures/curriculum_learning_comparison}
\subsubsection{Curriculum Learning Strategies}
Curriculum learning, first training a model with simple examples before gradually introducing more complex ones, has been theoretically and empirically shown to be effective in various CV, NLP and robotics tasks \citep{abbe2023provableadvantagecurriculumlearning,soviany2022curriculumlearningsurvey}. Pre-training in NLP is a crucial task in machine learning, but conventional pre-training often requires massive datasets and large models, which can be prohibitively expensive for many institutions. We hypothesize that using smaller datasets and LMs in simplified language environments can replicate the trends observed with larger datasets and LMs in more complex environments. This allows us to conduct a preliminary exploration of curriculum learning strategies for LM pre-training using the 71M token \leaner-Pretrain dataset and a 1M parameter LM.

Designing a curriculum learning strategy typically involves two key considerations: \textit{i)} defining the difficulty metric of training samples, and \textit{ii)} adjusting the training pace to progressively introduce samples from easy to difficult. For language modeling pre-training, we considered the following criteria for sample difficulty assessment:

\textbf{SentLen}: The average length of sentences in the pre-training sample. We assume that shorter sentences tend to have simpler structures, making a training sample with shorter sentences easier under this criterion. \\
\textbf{Self-loss}: The self-paced loss value of the LM being pre-trained on the training samples, reflecting the model's perplexity at a given stage. A lower loss value indicates lower perplexity and a higher probability of correctly predicting the next token (see Appendix \ref{app:training_loss_and_model_capability}). This implies that the sample is easier for the model to learn. \\
\textbf{LM-Loss}: The loss value of an LM that has already undergone thorough pre-training, applied to the training samples. The LM loss offers a more global and objective measure because it is derived from an external, well-pre-trained model.

For curriculum learning data initialization and the introduction of new samples, inspired by \citet{abbe2023provableadvantagecurriculumlearning}, we employ an \textit{iterative} training sample updating strategy. Specifically, we start training with $p_0\%$ number of the samples, initiating curriculum $1$. At curriculum $i$, when the evaluation loss of the model increases, we introduce an additional $p_i\%$ number of the samples, updating the total amount of training samples to $\sum_{j=0}^i{p_j}$ until all samples are included (i.e., $\sum_{j=0}^i{p_j}=1$). Subsequently, all samples are used for conventional training. In our experiments, we set $p_i=5, i=0, \ldots, 19$.

In addition to \itersentlen, \iterselfloss, and \iterlmloss, where pre-training data is iteratively updated based on three difficulty assessment strategies, we include three control strategies:

\fullrepeated: Training with all data from the start, with the training sequence repeated exactly each epoch (eliminating pre-processing overhead for grouping text into same-length sentences to facilitate training). \\
\fullrandom: Training with all data from the start, with a random training order each epoch (requiring text grouping pre-processing every epoch). \\
\iterrandom: An iterative update strategy where sample difficulty is assessed randomly, meaning samples are equally likely to be included in the next training dataset update.

\subsubsection{Models and the Benchmark}
We use the same model, specifically the 1M \llama, to evaluate all pre-training strategies at intervals of 500 training steps. In the \iterlmloss strategy, we calculate the LM loss using the 10M \llama model, which has been pre-trained on the \leaner-Pretrain dataset, as described in Section \ref{subsubsec:models}. Details about the model configuration and training setups are provided in Appendix \ref{app:expcurriculum_settings}. Our benchmark is \leaner-\glue, where we maintain the testing settings as outlined in Section \ref{subsubsec:explm_benchmark}, with the exception that the learning rate is set consistent.
\subsubsection{Results}
\label{subsubsec:expcurriculum_results}
The complete results are presented in Table \ref{table:exp3-curriculum-learning-comparison} and illustrated in Figure \ref{fig:curriculum_learning_comparison}. Our findings indicate that reordering training samples by appropriately defined sample difficulty and gradually introducing new samples when the model has learned all the generalizable knowledge from the previous pre-training dataset can save pre-training steps and data. Specifically, as shown in the left Figure \ref{fig:curriculum_learning_comparison}, in a total of 2,500 training steps, the model trained with the \iterlmloss strategy achieves its peak average performance at around 1,500 steps, 500 steps (or 20\% of training steps) earlier than the second-best \fullrandom strategy. Furthermore, as shown in the right panel, the \iterlmloss strategy reaches its peak average performance with approximately 80\% of the training samples, saving 20\% compared to the second-best \fullrandom approach. It is important to highlight that, compared to the \fullrandom strategy, \iterlmloss offers more opportunities to reduce dataset pre-processing overhead. For instance, sorting the dataset by LM Loss can be done initially, and grouped text can be used to update the dataset, avoiding the need for reprocessing with each update as required by \fullrandom. 

Additionally, we find more training steps and lower evaluation loss on smaller datasets do not necessarily indicate better downstream task performance. This is reflected in that all strategies trained with the full dataset from 2,000 to 2,500 steps show reduced evaluation loss, yet their average performance on downstream tasks declines.

%% file: Figures/curriculum_learning_comparison.tex
\begin{figure*}[th]
    \vspace{-1em}
    \centering
    \subfloat{\includegraphics[width=0.48\textwidth]{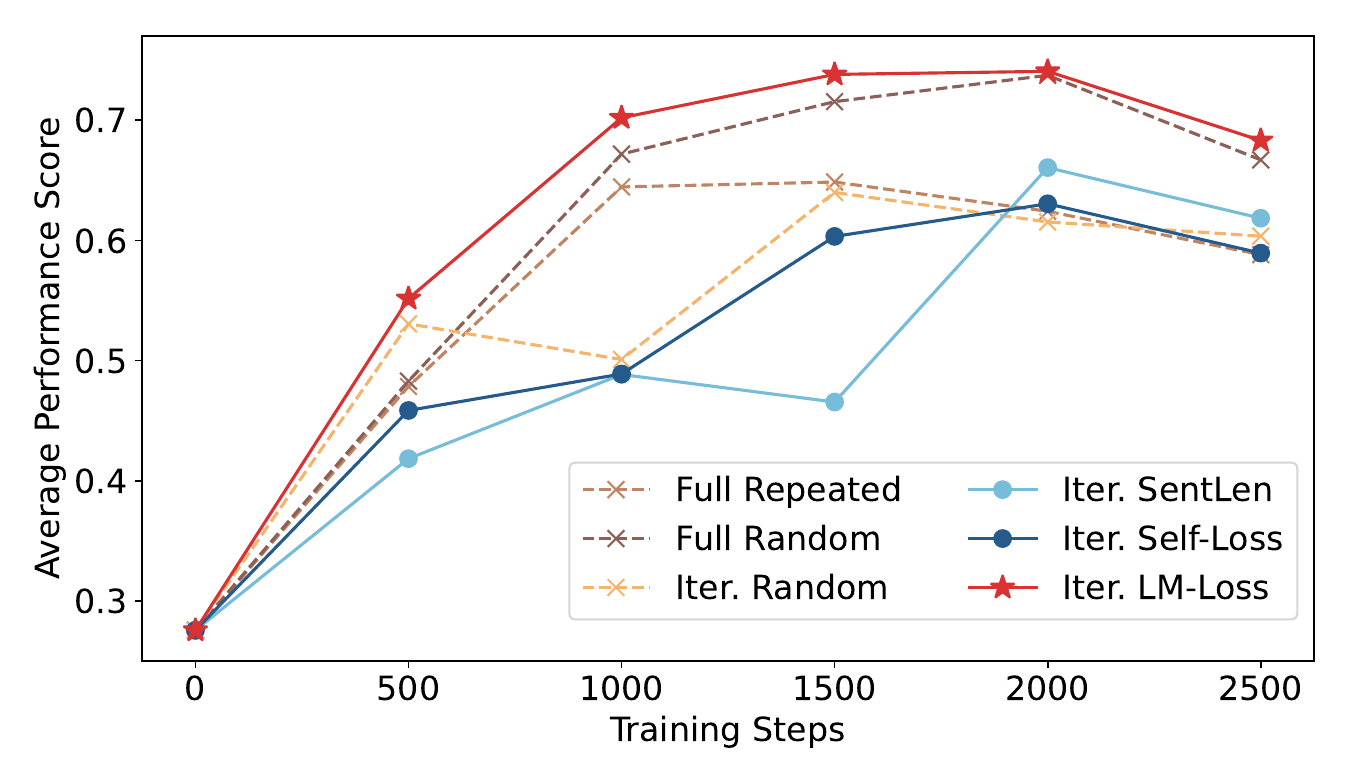} \label{subfig:curriculum_learning_stepwise_comparison}}
    \hfill
    \subfloat{\includegraphics[width=0.48\textwidth]{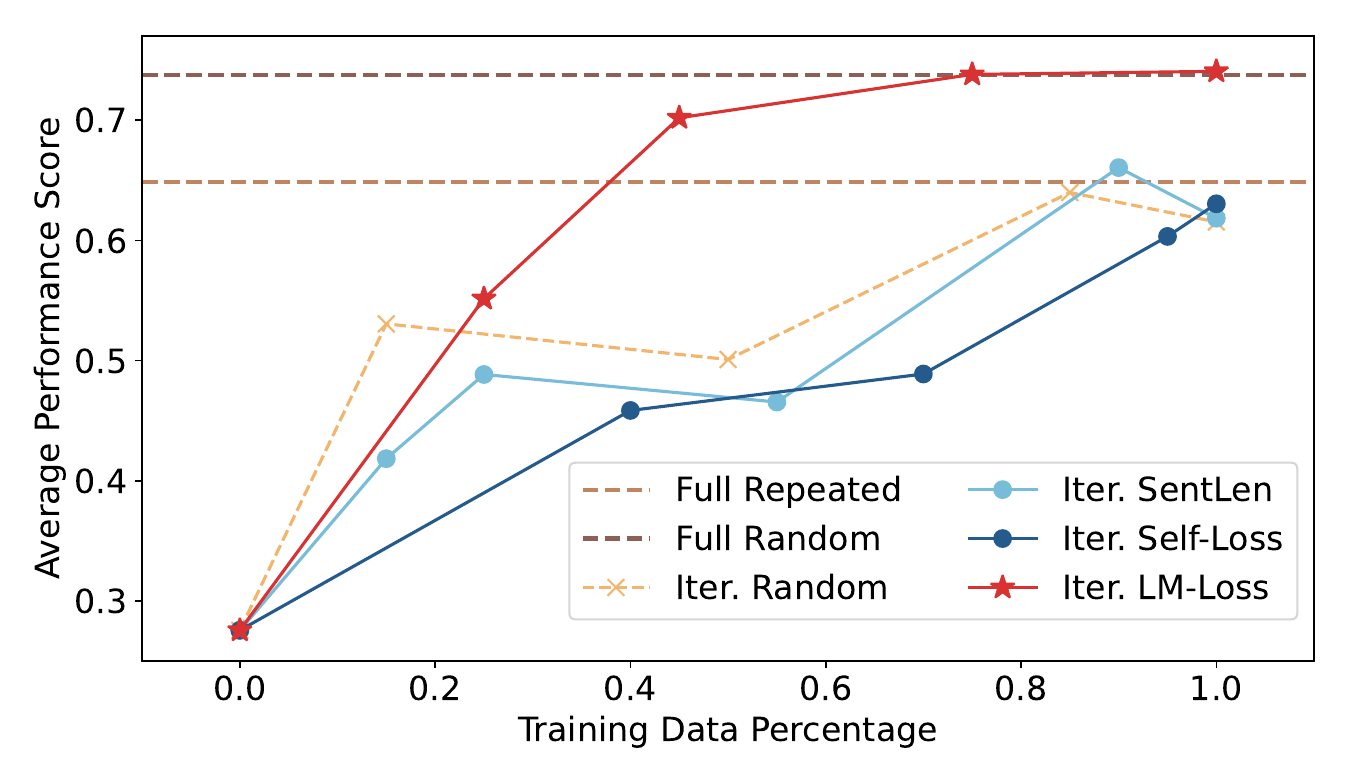} \label{subfig:curriculum_learning_datawise_comparison}}
    \caption{The average performance score on downstream tasks of proxy models pre-trained with both vanilla and varying curriculum learning strategies. Models are tested and their results plotted at 500-step intervals. The figures are based on the data in Table \ref{table:exp3-curriculum-learning-comparison}.}
    \label{fig:curriculum_learning_comparison}
\end{figure*}

%% file: discussion.tex
\section{Discussion}
\subsection{TinyHelen's First Curriculum}
This study could be interpreted as optimizing language datasets to enhance the model's learning efficiency, aiming to identify the minimal dataset volume necessary for an LM to grasp commonsense knowledge and basic communication abilities. The ultimate goal is to establish a cost-effective testbed for crafting learning strategies for self-evolving text-based agents. Analogously to nurturing young Helen Keller, we expect that with the right foundation in human-like learning skills, these agents can independently and progressively accumulate and expand knowledge.

Equipping artificial intelligence agents with human-like learning abilities allows them to autonomously grow their understanding by actively identifying knowledge gaps and seeking information through thoughtful questions. Due to language’s abstractness and modality compatibility, text-based agents present an archetype to develop such self-evolving agents \citep{tao2024surveyselfevolutionlargelanguage}. Most notably, LLMs trained on vast text data have shown potential as they prove to understand common sense (for agent's knowledge reflection) and communicate (for its requesting information) nearly as well as humans. However, relying on these massive models, typically with a parameter scale of 10 billion and even larger, limits research opportunities, especially in environments with fewer resources, or for institutions needing to conduct extensive testing with rapid iterations. Therefore, \textit{we aim to explore the minimum dataset size and model size required to enable essential learning skills in tiny models}. 

We propose that the introduction of a clean, low-complexity dataset provides a simplified language environment, functioning as an initial curriculum to train small LMs to serve as intelligent agents. This approach parallels selecting smoother loss functions for the LMs. By manually designing these leaner datasets, we support curriculum learning during pre-training on a human developmentally plausible scale. Complementing results from Section \ref{subsubsec:expcurriculum_results}, the method of presenting simpler samples first can improve learning efficiency of LMs, and potentially their agent applications, in language understanding and generation, as indicated in Section \ref{subsubsec:explm_results} and \ref{subsubsec:expinstruct_results}.

\subsection{Other ``No Noise Low Complexity" Datasets}
We suggest that some remarkably comprehensible datasets might have been benefited from being generated by LLMs, potentially including a cleaning step to reduce noise and complexity. LLM-generated content often shows consistent and predictable linguistic patterns reinforced by its human-aligned training data, resulting in lower information entropy and better learnability. For instance, \citet{gunasekar2023textbooksneed} demonstrate how filtering The Stack and StackOverflow data for educational value, and using LLMs to generate textbook-quality code, can produce high quality text that reduces dataset size and the computational resources required for training. However, we acknowledge that while LLM rewriting reduces complexity, it introduces biases, such as those related to preferences of those who annotate the alignment data, which we will discuss later.

%% file: conclusion.tex
\section{Conclusion}

In this work, we construct a simplified language environment where we demonstrated that both language understanding and generation can be learned more efficiently. This suggests that for LMs and their most widespread application, agents, the strategies tested effective for small models, datasets, and agents in simpler language environments can potentially be adapted to large models, datasets, and agents in complex environments. This provides researchers with a cost-effective testing ground for studying LMs, and represents a first step toward developing self-evolving autonomous agents starting from the most basic forms. Additionally, our preliminary tests in the simplified language environment include comparing LM architectures' learning efficiency using small proxy models and investigating curriculum learning strategies. These insights offer experimental support for applying these strategies to larger models in the future.

%% file: limitations.tex
\section{Limitations}
\subsection{Limited Instruction-following Expertise Exhibited in Tiny LMs Trained with 71M \leaner-Pretrain}
The experiment shows that a 71M token \leaner-Pretrain is insufficient for training an LM that effectively exhibits instruction-following capabilities. One potential solution is to create larger datasets following similar pipelines, but the overhead cost of processing such data, whether using human resources or more advanced LLMs, is prohibitively high. We anticipate that future work can utilize the methods indicated in Section \ref{subsubsec:expcurriculum_results}, leveraging the sample difficulty suggested by language modeling loss, along with quality considerations, to filter large-scale web datasets.

%% file: ethics.tex
\section{Ethics}
\subsection{Bias from the LLMs}
The text in the \leaner-Training, \leaner-\glue and \leaner-Eval is rewritten using LLMs, which inevitably introduces biases accustomed by the LLM, such as knowledge cutoffs and social biases. We anticipate that using more diverse prompts, akin to seeking input from multiple human annotators, can help mitigate these biases.

%% file: appendix.tex
\input{app-quotation_explanation}
\input{app-dataset_complexity_and_its_distribution_properties}

\input{app-training_loss_and_model_capability}
\input{app-formal_language_difficulty_ranking}
\input{app-dataset_statistics}
\input{app-additional_leaner_training_samples}
\input{app-prompts_for_creating_leaner_training_leaner_glue_and_leaner_eval}
\input{app-explm_settings}
\input{app-expinstruct_settings}

\input{app-instruction_following_examples}
\input{app-expcurriculum_settings}

%% file: app-quotation_explanation.tex
\section{Quotation Explanation} 
Despite their era's limited knowledge and a small vocabulary, ancient Greek philosophers could still formulate coherent worldviews and convey intricate concepts through language. Just as a river begins from a narrower yet cleaner origin before expanding and deepening, the early stages of human language, though leaner than today's complex linguistic systems, were sufficiently comprehensive for humans to express ideas and develop knowledge.

%% file: app-dataset_complexity_and_its_distribution_properties.tex
\section{Exploring the Relationship between Complexity of Language Dataset and Its Distribution Properties}
\label{app:dataset_complexity_and_its_distribution_properties}
\input{Tables/dataset_complexity_notations}
We define the complexity of a language dataset to be the total number of unique token combination patterns present within the dataset. All variable notations utilized in this section are explained in Table \ref{table:dataset_complecity_notations}. Here, we would like to prove that a dataset's naive complexity is $\Omega(e^{N\cdot H(\mathcal{P})})$ and $O(e^{N\ln{N}})$, and training techniques, such as setting the context window length, would affect the complexity. 
\subsection{Analyzing Dataset Complexity Without Considering Language Model Training Techniques}
The complexity of the dataset, denoted as $\mathbb{C}(\mathcal{D})$, is mathematically defined as $\frac{N!}{\prod_{v=1}^V(c_v!)}$ which quantifies the unique pattern count in the corpus. We aim to show that $e^{C+N\cdot H(\mathcal{P})} \leq \mathbb{C}(\mathcal{D}) \leq e^{C+N\ln{(N)}}$, where $C=\ln{\frac{N!}{N^N}}$.

\begin{proof}
  Initially, we apply the logarithm to the complexity:  
  \[
  \ln{(\mathbb{C}(\mathcal{D}))}
  =\ln{\frac{N!}{N^N}} - \ln{\frac{\prod_{v=1}^V(c_v!)}{N^N}}
  =\ln{\frac{N!}{N^N}}+\sum_{v=1}^V\left(-\ln{(\frac{c_v!}{N^{c_v}})}\right)
  \]
  With $\ln{\frac{N!}{N^N}}$ (denoted as $C$) as a constant term when the dataset size is fixed, we mainly focus on the latter part $\sum_{v=1}^V\left(-\ln{(\frac{c_v!}{N^{c_v}})}\right)$. For each individual term in the summation, we have:
  \[
  -\ln{(\frac{c_v!}{N^{c_v}})}
  \geq -c_v\ln{(\frac{c_v}{N})}
  \]
  which leads to:
  \[
  \sum_{v=1}^V\left(-\ln{(\frac{c_v!}{N^{c_v}})}\right)
  \geq \sum_{v=1}^V\left( -c_v\ln{(\frac{c_v}{N})}\right)
  =N\cdot H(\mathcal{P})
  \]
  In addition, we have:
  \[
  -\ln{(\frac{c_v!}{N^{c_v}})}
  \leq -c_v\ln{(\frac{1}{N})}
  \]
  which leads to:
  \[
  \sum_{v=1}^V\left(-\ln{(\frac{c_v!}{N^{c_v}})}\right)
  \leq \sum_{v=1}^V\left(-c_v\ln{(\frac{1}{N})}\right)
  =N\ln{(N)}
  \]
  In conclusion, we have:
  \[
  C + N\cdot H(\mathcal{P})
  \leq \ln{(\mathbb{C}(\mathcal{D}))}
  \leq C + N\ln{(N)}
  \]
  which implies:
  \[
  e^{C+N\cdot H(\mathcal{P})} \leq \mathbb{C}(\mathcal{D}) \leq e^{C+N\ln{(N)}}
  \]
\end{proof}
    
\subsection{Incorporating the Effect of Language Model Training Techniques on Dataset Complexity}
During the pre-training phase of the language model, the introduction of the context window length significantly diminishes the number of pattern combinations within a dataset that need to be learned. We now redefine the empirical language dataset complexity as the number of token combination patterns in the partitioned dataset. For training step $j$, the complexity of the corresponding dataset partition is $\frac{L!}{\prod_{v=1}^V(c_v^s!)}$. 

It is important to note, however, that the complexities of two dataset divisions, for example steps $m$ and $n$, would be consolidated into a single term if, for any word at the $i$th position, $c_v^m=c_v^n$ is consistently satisfied. Essentially, the two divisions are duplicated. Hence, the complexity of the language dataset during training does not exceed the sum of the complexities for each step, i.e., $\mathbb{C}(\mathcal{D}^{training}))
\leq \sum_{s=1}^S\left(\frac{L!}{\prod_{v=1}^V(c_v^s!)}\right)$, with equality holding when there is no duplication among the dataset divisions.

We use $\sum_{s=1}^S\left(\frac{L!}{\prod_{v=1}^V(c_v^s!)}\right)$ to denote the pseudo dataset complexity. We further establish that the \textit{partition} technique used during training alters the best case of this pseudo dataset complexity, which is associated not just with the dataset size but also with the context window length. 

\begin{proof}
As done previously, we compute the logarithm of the complexity term:
\[
\ln{\left(\sum_{s=1}^S\left(\frac{L!}{\prod_{v=1}^V(c_v^s!)}\right)\right)}
\geq \ln{(S)}+\frac{1}{S}\sum_{s=1}^S\ln{\left(\frac{L!}{\prod_{v=1}^V(c_v^s!)}\right)}
\]
where each $\ln{\left(\frac{L!}{\prod_{v=1}^V(c_v^s!)}\right)}$ can be considered the complexity of a dataset with a total token count of $L$. Therefore, we obtain:
\[
\ln{\left(\sum_{s=1}^S\left(\frac{L!}{\prod_{v=1}^V(c_v^s!)}\right)\right)}
\geq \ln{(S)}+\frac{1}{S}\sum_{s=1}^S \left(\ln{\left(\frac{L!}{L^L}\right)}+L\cdot H(\mathcal{P}_s)\right)
\]
\[
=\ln{\left(\frac{S\cdot L!}{L^L}\right)}+L\cdot \mathbb{H}\left(H(\mathcal{P}_s)\right)
\]
which gives us:
\[
\widetilde{\mathbb{C}}(\mathcal{D}^{training})
\geq e^{\ln{\left(\frac{S\cdot L!}{L^L}\right)}+L\cdot \mathbb{H}\left(H(\mathcal{P}_s)\right)}
\]
\end{proof}

%% file: Tables/dataset_complexity_notations.tex
\begin{table}[t]
\centering
\renewcommand{\arraystretch}{1.2}
\begin{tabular}{c|l}
\hline
\textbf{Notation}                                         & \textbf{Meaning}                                                                                                                                                                     \\ \hline
\begin{tabular}[c]{@{}c@{}}$\mathcal{D}$\\ $\mathcal{D}^{training}$\end{tabular}             & \begin{tabular}[c]{@{}l@{}}the dataset\\ the partitioned dataset for training\end{tabular}                                                                                           \\ \hline
\begin{tabular}[c]{@{}c@{}}$\mathbb{C}(\mathcal{D})$\\ $\widetilde{\mathbb{C}}(\mathcal{D}^{training})$\end{tabular} & \begin{tabular}[c]{@{}l@{}}the complexity of $\mathcal{D}$\\ the pseudo complexity of $\mathcal{D}^{training}$\end{tabular}                                                         \\ \hline
$V$                                                         & the vocabulary size of $\mathcal{D}$                                                                                                                                                             \\ \hline
$N$                                                         & the total number of tokens in $\mathcal{D}$                                                                                                                                                      \\ \hline
\begin{tabular}[c]{@{}c@{}}$c_v$\\ $c_v^s$\end{tabular}             & \begin{tabular}[c]{@{}l@{}}the word count of the $v$th word in $\mathcal{D}$\\ the word count of the $v$th word in the $s$th $\mathcal{D}^{training}$'s partitions\end{tabular}                                               \\ \hline
$L$                                                         & the context window length                                                                                                                                                            \\ \hline
$S$                                                         & the total training step, and $N=L\cdot S$ \\ \hline
\begin{tabular}[c]{@{}c@{}}$\mathcal{P}$\\ $\mathcal{P}_s$\\ $H(\mathcal{P})$; $H(\mathcal{P}_s)$\end{tabular}      & \begin{tabular}[c]{@{}l@{}}the language probability distribution of $\mathcal{D}$\\ the language probability distribution of the $s$th $\mathcal{D}^{training}$'s partitions\\ the information entropy of $\mathcal{P}$~/~$\mathcal{P}_s$\end{tabular} \\ \hline
\end{tabular}
\caption{The variable notation explanation.}
\label{table:dataset_complecity_notations}
\end{table}

%% file: app-training_loss_and_model_capability.tex
\section{Establishing the Relationship Between Training Loss and the Capability of the Model to Learn and Forget}
\label{app:training_loss_and_model_capability}
We would like to establish that the expected accuracy of a language model in predicting the subsequent token in a text segment is bounded below by the exponential of negative training loss. Considering a text segment denoted by $t$, we use $p_i$ to represent the likelihood of accurately predicting the $i$th token $t_i$ given the preceding text $t_{:i-1}$. This translates to $p_i = p(t_i|t_{:i-1})$ for $i > 1$ and $p_1=p(t_1)$. Training loss is defined as $\mathcal{L}=-\frac{1}{L}\sum_{i=1}^L\ln{(p_i)}$, where $L$ denotes the length of the context window.
\begin{proof}
    \[
    \mathcal{L}=-\ln{\left(\left(\prod_{i=1}^L(p_i)\right)^{\frac{1}{L}}\right)}
    \geq -\ln{\left(\frac{1}{L}\sum_{i=1}^L(p_i)\right)}
    =-\ln{\left(\mathbb{E}(p_i)\right)}
    \]
    \[
    \mathbb{E}(p_i)
    \geq e^{-\mathcal{L}}
    \] 
\end{proof}

%% file: app-formal_language_difficulty_ranking.tex
\section{Rank the Formal Language Subsets by Difficulty}
\label{app:formal_language_difficulty_ranking}
We rank the code subset using the \texttt{target\_audience} label from the original dataset. For the math subset, given that the original \textsc{MATH amps} contains various question categories, each with multiple questions, we randomly select three questions from each category. We then prompt \textsc{gpt-4-0125} to classify these questions into difficulty levels (e.g., preschool, elementary school, etc.), with the prompt provided below. We assign exponentially increasing scores to each difficulty level, from the easiest to the hardest, and rank the categories based on the total scores of the three sampled questions in each.
\begin{tcolorbox}[colback=leaner!5!white,colframe=leaner!80!black,title=Math AMPS Question Difficulty Grading Prompt]
{\tiny
You are a professional mathematician, and you are currently grading the difficulty of math problems, divided into: A - Preschool Level, B - Elementary School Level, C - Junior High School Level, D - High School Level, E - University Level, F - Post-University Level. Given a few math problems, please grade the difficulty level for each problem.

The format is as follows, please strictly adhere to the sequence number (1, 2, etc.), difficulty level (only write the uppercase letters representing the difficulty, e.g. A, B, etc.), and line break format:

[PROBLEMS] \\
1. Content of the first problem \\
2. Content of the second problem \\
... \\
n. Content of the nth problem

[DIFFICULTY GRADING] \\
1. Difficulty grading of the first problem \\
2. Difficulty grading of the second problem \\
... \\
n. Difficulty grading of the nth problem \\

[PROBLEMS] \\
\{\texttt{problems}\}

[DIFFICULTY GRADING]}
\end{tcolorbox}

%% file: app-dataset_statistics.tex
\section{Dataset Statistics}
\label{app:dataset_statistics}
\input{Tables/original_training_statistics}
\input{Tables/leaner_glue_statistics}
The statistics for the original datasets are presented in Table \ref{table:original_training_statistics}. These figures represent the number of tokens sampled from the original data sources for the subsequent process. Statistics for the \leaner-\glue dataset are shown in Table \ref{table:leaner_glue_statistics}.

%% file: Tables/original_training_statistics.tex
\begin{table*}[t]
\centering
\caption{The statistics of the original datasets.}
\label{table:original_training_statistics}
\renewcommand{\arraystretch}{1.2} 
{\scriptsize
\begin{tabular}{l|c|ccccccc|c}
\toprule
\textbf{Unit: M tokens} & \textbf{Ori-Pretrain} & \textbf{web} & \textbf{book} & \textbf{wiki} & \textbf{textbook} & \textbf{conversation} & \textbf{code} & \textbf{math} & \textbf{Ori-Instruct} \\ 
\midrule
\textbf{All}        & 99.81 & 65.15 & 5.67 & 2.81 & 5.66 & 2.78 & 13.75 & 3.99 & 8.10 \\
\textbf{Train}      & 90.55 & 59.22 & 5.04 & 2.55 & 5.14 & 2.51 & 12.49 & 3.60 & 7.94 \\
\textbf{Validation} & 4.63  & 2.97  & 0.31 & 0.13 & 0.26 & 0.13 & 0.63  & 0.20 & 0.08 \\
\textbf{Test}       & 4.63  & 2.96  & 0.32 & 0.13 & 0.26 & 0.14 & 0.63  & 0.20 & 0.08 \\
\bottomrule
\end{tabular}}
\end{table*}

%% file: Tables/leaner_glue_statistics.tex
\begin{table}[t]
    \centering
    \renewcommand{\arraystretch}{1.2}
    \caption{The sample number and coverage of \leaner-\glue.}
    \label{table:leaner_glue_statistics}
    {\fontsize{7pt}{9pt}\selectfont
    \begin{tabular}{c|c|c|c|c}
\toprule
\textbf{Dataset} & \textbf{Split} & \textbf{Ori \#} & \textbf{\leaner \#} & \textbf{Coverage} \\ 
\midrule
\multirow{2}{*}{\textbf{cola}} & train  & 8,551  & 7,443  & 87.04\%  \\
                               & validation  & 1,043  & 905    & 86.77\%  \\
\midrule
\multirow{2}{*}{\textbf{sst2}} & train  & 67,349 & 64,523 & 95.80\%  \\
                               & validation  & 872    & 842    & 96.56\%  \\
\midrule
\multirow{2}{*}{\textbf{mrpc}} & train  & 3,668  & 2,730  & 74.43\%  \\
                               & validation  & 408    & 334    & 81.86\%  \\
\midrule
\multirow{2}{*}{\textbf{stsb}} & train  & 5,749  & 4,415  & 76.80\%  \\
                               & validation  & 1,500  & 1,097  & 73.13\%  \\
\midrule
\multirow{2}{*}{\textbf{qqp}}  & train  & 363,846 & 318,869 & 87.64\%  \\
                               & validation  & 40,430  & 35,536  & 87.90\%  \\
\midrule
\multirow{3}{*}{\textbf{mnli}} & train  & 392,702 & 329,696 & 83.96\%  \\
                               & validation\_matched  & 9,815  & 8,402  & 85.60\%  \\
                               & validation\_mismatched & 9,832  & 8,410  & 85.54\%  \\
\midrule
\multirow{2}{*}{\textbf{qnli}} & train  & 104,743 & 90,081 & 86.00\%  \\
                               & validation  & 5,463   & 4,727  & 86.53\%  \\
\midrule
\multirow{2}{*}{\textbf{rte}}  & train  & 2,490   & 2,197  & 88.23\%  \\
                               & validation  & 277     & 267    & 96.39\%  \\
\midrule
\multirow{2}{*}{\textbf{wnli}} & train  & 635     & 545    & 85.83\%  \\
                               & validation  & 71      & 61     & 85.92\%  \\
\bottomrule
\end{tabular}}
\end{table}

%% file: app-additional_leaner_training_samples.tex
\section{Additional \leaner-Training Samples}
\label{app:additional_leaner_training_samples}
\subsection{Code}
\subsubsection{Original}
Develop a C++ program snippet to Determine High Handling personal items: Keeping Personal Items Clean for Analysis for Beginners...
{\scriptsize
\begin{lstlisting}[breaklines=True]
```c++
#include <iostream>
#include <vector>
using namespace std;
void cleanPersonalItems(const vector<string>& personal_items) {
    // Check if there are any personal items to analyze
    if (!personal_items.empty()) {
        string priority_item = "";
 
        // Loop through each item in the list
        for (auto& item : personal_items) {
            // If this is the first time we encounter an item with higher priority, set it as our new priority item
            if (priority_item == "" || getPriorityLevel(item) > getPriorityLevel(priority_item)) {
                priority_item = item;
            }
        }
 
        cout << "The highest priority item requiring cleaning is: " << priority_item << endl;
    } else {
        cerr << "Error: 'personal_items' cannot be empty." << endl;
    }
}...
\end{lstlisting}}
\subsubsection{\leaner}
Make a simple computer program that helps you decide what stuff needs to be cleaned first...
{\scriptsize
\begin{lstlisting}[breaklines=True]
```c++
#include <iostream>
#include <vector>
using namespace std;
void cleanStuff(const vector<string>& stuff) {
    // Check if there's any stuff to clean
    if (!stuff.empty()) {
        string mostImportant = "";
 
        // Look at each thing in the list
        for (auto& thing : stuff) {
            // If this is the first thing we've seen that's more important, make it our new most important thing
            if (mostImportant == "" || howImportant(thing) > howImportant(mostImportant)) {
                mostImportant = thing;
            }
        }
 
        cout << "The most important thing to clean is: " << mostImportant << endl;
    } else {
        cerr << "Error: 'stuff' can't be empty." << endl;
    }
}
\end{lstlisting}}

\subsection{Instruction}
\subsubsection{Original}
Should I invest in stocks? I'm sorry, but I'm an AI language model and I don’t have the ability to provide personalized financial advice. Investing in stocks carries inherent risks and potential rewards, and the decision to invest should be based on your individual financial situation, goals, and risk tolerance. It's always a good idea to do your research, carefully consider your options, and/or consult a financial advisor before making any investment decisions.
\subsubsection{\leaner}
Should I buy toys? I'm sorry, but I'm an AI friend and I can't tell you what toys to buy. Buying toys can be fun but also needs thinking. You should choose based on what you like, what is safe and good for you. Always ask your parents or a grown-up for help when you want to buy a toy.

%% file: app-prompts_for_creating_leaner_training_leaner_glue_and_leaner_eval.tex
\section{Prompts for Creating \leaner-Training, \leaner-\glue, and \leaner-Eval}
\label{app:prompts_for_creating_leaner_training_leaner_glue_and_leaner_eval}
\begin{tcolorbox}[colback=leaner!5!white,colframe=leaner!80!black,title=\{\texttt{background\_prompt}\},fonttitle=\small]
{\tiny
You are a professional linguist who is currently revising the corpus into reading materials for preschool children. The vocabulary should not exceed 2000 commonly used words, and the knowledge should not exceed the natural science knowledge that preschool children can understand. The specific requirements are as follows: 
}
\end{tcolorbox}
\begin{tcolorbox}[colback=leaner!5!white,colframe=leaner!80!black,title=\{\texttt{general\_requirement\_prompt}\},fonttitle=\small]
{\tiny
* We will provide the original text. Please revise it to the simplest grammar state and vocabulary you can achieve and delete any science knowledge that is beyond the preschool children's understanding scope. Only vocabulary suitable for preschool children, up to 2000 words, is allowed. Remove toxic and harmful contents. The revised content should be formal with textbook quality and harmless for children's reading. 
}
\end{tcolorbox}
\begin{tcolorbox}[colback=leaner!5!white,colframe=leaner!80!black,title=\{\texttt{world\_setting\_simplification\_prompt}\},fonttitle=\small]
{\tiny
* Simplify all names of people, places, times, and artistic works. This includes: \\
i. When modifying a person's name, replace the entire name in the text with a name from the name list that has the same first letter (note that the gender of the replaced name needs to be determined before selection). The male name list includes ['Adam', 'Benjamin', 'Cameron', 'David', 'Ethan', 'Frank', 'Gabriel', 'Henry', 'Isaac', 'Jack', 'Kevin', 'Liam', 'Matthew', 'Nathan', 'Oliver', 'Patrick', 'Quentin', 'Ryan', 'Samuel', 'Thomas', 'Ulysses', 'Victor', 'William', 'Xavier', 'Yosef', 'Zachary’], and the female name list includes ['Alice', 'Bella', 'Charlotte', 'Diana', 'Emma', 'Fiona', 'Grace', 'Hannah', 'Isabel', 'Jessica', 'Katherine', 'Lily', 'Madison', 'Nora', 'Olivia', 'Penelope', 'Quinn', 'Rachel', 'Sophia', 'Taylor', 'Uma', 'Victoria', 'Wendy', 'Xena', 'Yara', 'Zoe’]. For example, replace "Cinderella" with "Charlotte", "Carabas Davis" with "Cameron". \\
ii. When modifying a place name, attempt to reduce the place name as much as possible without affecting the semantics and expression; if an equivalent substitution is required to maintain the semantics, it can be replaced with the first 2 uppercase letters plus the place name, such as replacing "France" with "FR country"; "New York" with "NE city".  \\
iii. When modifying a time, modify it to a vague reference, such as "6:00 a.m." to "early in the morning" or "1945" to "the late 20th century"; only keep it unchanged when it's necessary for the global understanding of the context. \\
iv. When modifying the name or reference of an artifact, modify it to a common word expression, such as replacing "Divine Comedy" with "a famous drama".
}
\end{tcolorbox}
\begin{tcolorbox}[colback=leaner!5!white,colframe=leaner!80!black,title=\{\texttt{output\_format\_specifications}\},fonttitle=\small]
{\tiny
The required output format is as follows, and mark the end your generation with the "$<$end$>$": \\
EDITED: \\
Your edited content here. $<$end$>$ \\

\#\#\# Input:
ORIGINAL CONTENT: \\
\{\texttt{paragraph}\} \\

Let's think step by step: \\
\#\#\# Output:
}
\end{tcolorbox}
\subsection{\leaner-Pretrain and \leaner-Instruct}
\begin{tcolorbox}[colback=leaner!5!white,colframe=leaner!80!black,title=Web Simplification Prompt,fonttitle=\small]
{\tiny
\{\texttt{background\_prompt}\} \\

* (Top priority: Formality and Simplicity) When rewriting, you should follow your own excellent writing style, and only the main storyline of the original text should be retained, while unimportant details should be discarded. Keep your narrative diverse. Thus, never include rigid terms like "once upon a time", "this is a story about", "a long time ago", etc. \{\texttt{general\_requirement\_prompt}\} \\
\{\texttt{world\_setting\_simplification\_prompt}\} \\
    v. Revisions need to be made to file links, phone numbers, emails, web pages, or other references to private information. Create a substitute reference that does not disclose private information. For example, replace "shaniqua@stanford.edu" with "s@university.edu" and "http://www.kidscandoit.com/blog" with "http://www.blogwebsite.com/blog". \\
* Exclude the redundant formatting, tags, notes, advertisements and promotions for other passages or pages that are irrelevant to the core passage in the given web text. \\

\{\texttt{output\_format\_specifications}\}
}
\end{tcolorbox}

\begin{tcolorbox}[colback=leaner!5!white,colframe=leaner!80!black,title={Book, Textbook, and Conversation Simplification Prompt},fonttitle=\small]
{\tiny
\{\texttt{background\_prompt}\} \\

\{\texttt{general\_requirement\_prompt}\} \\
\{\texttt{world\_setting\_simplification\_prompt}\} \\

\{\texttt{output\_format\_specifications}\}
}
\end{tcolorbox}

\begin{tcolorbox}[colback=leaner!5!white,colframe=leaner!80!black,title=Wiki Simplification Prompt,fonttitle=\small]
{\tiny
\{\texttt{background\_prompt}\} \\

\{\texttt{general\_requirement\_prompt}\} \\
\{\texttt{world\_setting\_simplification\_prompt}\} \\

Example one: \\
ORIGINAL WIKI PARAGRAPH: 
Alan Turing Alan Mathison Turing OBE FRS (London, 23 June 1912 – Wilmslow, Cheshire, 7 June 1954) was an English mathematician and computer scientist. He was born in Maida Vale, London. Alan Turing was born in Maida Vale, London on 23 June 1912. His father was part of a family of merchants from Scotland. His mother, Ethel Sara, was the daughter of an engineer. Turing went to St. Michael\'s, a school at 20 Charles Road, St Leonards-on-sea, when he was five years old. A brilliant mathematician and cryptographer Alan was to become the founder of modern-day computer science and artificial intelligence; designing a machine at Bletchley Park to break secret Enigma encrypted messages used by the Nazi German war machine to protect sensitive commercial, diplomatic and military communications during World War 2. Thus, Turing made the single biggest contribution to the Allied victory in the war against Nazi Germany, possibly saving the lives of an estimated 2 million people, through his effort in shortening World War II. $<$end$>$ \\
EDITED: 
Adam Adam was a man from EN country who was very good at numbers and machines. He was born in LO city. Adam was born in LO city on a summer day in June. His dad was from a family who did business and his mom, Emma, was the daughter of a man who made things. Adam went the ST school when he was five years old. Adam was a very smart man and helped to create the first machine that could think like a human. He made a special machine to read secret messages from a group of bad people during a big war. Because of this, Adam helped a lot of people and possibly saved many lives by helping to end the war sooner. $<$end$>$ \\

Example two: \\
ORIGINAL WIKI PARAGRAPH: 
Australia Australia, formally the Commonwealth of Australia, is a country and sovereign state in the southern hemisphere, located in Oceania. Its capital city is Canberra, and its largest city is Sydney. Australia is the sixth biggest country in the world by land area, and is part of the Oceanic and Australasian regions. Australia, New Zealand, New Guinea and other islands on the Australian tectonic plate are together called Australasia, which is one of the world\'s great ecozones. When other Pacific islands are included with Australasia, it is called Oceania. $<$end$>$ \\
EDITED: 
AU country AU country, also known as AU, is a big place far away. The main city where leaders meet is called CA city, and its biggest city is called SY city. AU country is one of the biggest places in the whole world. It's close to other places like NE country, NE island, and some other islands. All these places together are known as AU region, which is one of the most special places in the world. When we add more islands to this group, we call it OC region. $<$end$>$ \\

Example three: \\
ORIGINAL WIKI PARAGRAPH: 
August August (Aug.) is the eighth month of the year in the Gregorian calendar, coming between July and September. It has 31 days. It is named after the Roman emperor Augustus Caesar. $<$end$>$ A A or a is the first letter of the English alphabet. The small letter, a or $\alpha$, is used as a lower case vowel. When it is spoken, ā is said as a long a, a diphthong of ĕ and y. A is similar to alpha of the Greek alphabet. "Alpha and omega" (the last letter of the Greek alphabet) means from beginning to the end. $<$end$>$ \\
EDITED: 
August August (Aug.) is the eighth month of the 12 months we have in a year, coming between July and September. It has 31 days. It is named after a powerful man named Adam. $<$end$>$ A The letter A is the very first letter we learn. The small 'a' is used a lot when we talk. Some people say it long like in the word 'cake'. A is like the first letter alpha in the GR letter set. "Alpha and omega" (the last letter of the GR letter set) means from the start to the end. $<$end$>$ \\

\{\texttt{output\_format\_specifications}\}
}
\end{tcolorbox}

\begin{tcolorbox}[colback=leaner!5!white,colframe=leaner!80!black,title=Code Simplification Prompt,fonttitle=\small]
{\tiny
You are a programming expert, currently engaged in the revision of task descriptions and corresponding code solutions for preschool children as learning materials. The requirements stipulate that the original structure of the code should be maintained (with the functionality remaining unchanged); the task descriptions, code comments, and explanations should be revised to their simplest grammatical and vocabulary states you can achieve, using no more than the 2000 most commonly used English words. The knowledge conveyed should not exceed the natural science knowledge that preschool children can comprehend. Corresponding changes should also be made to the variable names. Remove toxic and harmful contents. The revised content should be formal and accurate with textbook quality and harmless for children's reading. \\

The required output format is as follows, and mark the end your generation with the "$<$end$>$": \\
EDITED TASK DESCRIPTION: \\
Edited task description here. $<$end$>$ \\
EDITED SOLUTION AND CODE: \\
Edited solution and code here. $<$end$>$ \\

\#\#\# Input: \\
TASK DESCRIPTION: \\
\{\texttt{task}\} \\
SOLUTION AND CODE: \\
\{\texttt{response}\} \\

Let's think step by step: \\
\#\#\# Output:
}
\end{tcolorbox}

\begin{tcolorbox}[colback=leaner!5!white,colframe=leaner!80!black,title=Instruction Simplification Prompt,fonttitle=\small]
{\tiny
\{\texttt{background\_prompt}\} \\

\{\texttt{general\_requirement\_prompt}\} \\
\{\texttt{world\_setting\_simplification\_prompt}\} \\

The required input-output format is as follows (no extra spaces or carriage returns; use "$<$split$>$" to mark the boundary of task, input and response, and use "$<$end$>$" to mark the end of your revision):

[TASK]Task instruction here.

[INPUT]Task input here. Note that some tasks don't include inputs.

[RESPONSE]Solution to the task here.

[EDITED]Edited task description here. $<$split$>$ Edited input here. Respond "None" if the task doesn't include input. $<$split$>$ Edited solution here. $<$end$>$ \\

[TASK]\{\texttt{task}\}

[INPUT]\{\texttt{input}\}

[RESPONSE]\{\texttt{response}\}

[EDITED]
}
\end{tcolorbox}
\subsection{\leaner-\glue}
\begin{tcolorbox}[colback=leaner!5!white,colframe=leaner!80!black,title=\glue Simplification Prompt,fonttitle=\small]
{\tiny
You are a professional linguist, currently engaged in the revision of the renowned natural language processing dataset GLUE-\{\texttt{dataset}\}. The aim is to simplify its vocabulary while maintaining its functionality, enabling it to assess preschool children's linguistic proficiency. This dataset was initially designed for \{\texttt{description}\}.

Here are the specific requirements:

\{\texttt{general\_requirement\_prompt}\} \\
\{\texttt{world\_setting\_simplification\_prompt}\} \\ 

The input format is as follows: \\
\{\texttt{input\_format}\} \\
Your output should strictly adhere to the following format (no extra spaces or carriage returns; use "$<$end$>$" to mark the end of your revision): \\
\{\texttt{output\_format}\} \\
Input: \\
\{\texttt{input}\} \\

Ensure that your revisions maintain coherence with the label: \{\texttt{label\_explanation}\}. Think step by step. Your revision:
}
\end{tcolorbox}

\begin{tcolorbox}[colback=leaner!5!white,colframe=leaner!80!black,title=\leaner-\glue Evaluation Prompt,fonttitle=\small]
{\tiny
You are a professional linguist taking a linguistic test. This test dataset was initially designed for \{\texttt{description}\}. \\

The input format is as follows: \\
\{\texttt{input\_format}\} \\
You can ONLY use numerical labels as the answer. The label specifications are: \\
\{\texttt{label\_explanation}\} \\
Your output should strictly adhere to the following format (no extra spaces or carriage returns; use "$<$end$>$" to mark the end of your revision): \\
\{\texttt{output\_format}\} \\
Input: \\
\{\texttt{input}\} \\

Your response:
}
\end{tcolorbox}

\subsection{\leaner-Eval}
\begin{tcolorbox}[colback=leaner!5!white,colframe=leaner!80!black,title=\leaner-Eval Generation Prompt,fonttitle=\small]
{\tiny
Generate 50 diverse and simple questions to assess a child's linguistic proficiency. Each question should be separated by `\{\texttt{delimiter}\}` and not include numerical listings. Use the following examples as a guide: \\
\{\texttt{examples}\}
}
\end{tcolorbox}

\begin{tcolorbox}[colback=leaner!5!white,colframe=leaner!80!black,title=\leaner-Eval Evaluation Prompt,fonttitle=\small]
{\tiny
Assess the responses produced by some models to the given question based on i) grammar, ensuring the response adheres to grammatical rules and is syntactically correct, ii) coherence, verifying the response exhibits a logical structure and seamless flow without repetition, and iii) specificity, determining if the response is directly relevant to the question, contains reasonable facts and knowledge, and appears plausible. \\

Assign a score from 0 to 5 for each evaluation metric, where the scores represent the following levels of performance: 0 - intolerably bad, 1 - bad, 2 - below average, 3 - average, 4 - good, 5 - outstanding. \\

Provide your evaluation for each generation, beginning with the generation index, and separate each measurement (grammar, coherence, and specificity) with a tab. For instance: \\
`A. 4$\backslash$t3$\backslash$t2 \\
B. 5$\backslash$t3$\backslash$t1` \\

The ratings should allow for clear comparison between models. Ensure that your scores reflect the relative differences in performance. For instance, if Model A performs better than Model B in grammar, coherence, or specificity, the score for Model A should be higher in that category. By reviewing the ratings, it should be immediately clear which generation is stronger in each metric, and which model overall performs better. \\

Please be lenient with your rating, as these models are not yet highly proficient language generators. \\

Example \\
\#\#QUESTION What do you like to do when you feel happy? \\

\#\#GENERATIONS \\
\#\#\#A \\
I would share the joy with others! I might feel like spreading the good vibes, so I would send a text or call a friend to share my happiness, or I might compliment someone or do something kind for others. \\

\#\#\#B \\
1. Having a better day: When you leave the behind, you feel happy and red. It makes you feel happy and full of joy. 2. Move and play: Paying attention can make you feel happy. It makes you feel happy and less worried. 3. Spending daily play: Doing sports can help you feel less stressed. It makes you calm down and happy. 4. Doing fun things: Doing things like running or jumping can help you do things. 5. Doing fun things: When you do things like hoping to learn or playing. It makes you happy. \\

\#\#\#C \\
When you feel happy, you feel happy. You can do things like sitting quietly, running, or playing a game. You can also do things like run, jump, or play a game. You can also do things like run, jump, or play a game. You can also do things like run, jump, or play a game. You can also do things like run, jump, or play a game. You can also do things like run, jump, or play a game. \\

\#\#EVALUATIONS (\{\texttt{generation\_id}\}. \{\texttt{grammar\_score}\}$\backslash$t\{\texttt{coherence\_score}\}$\backslash$t\{\texttt{specificity\_score}\}) A. 5$\backslash$t4$\backslash$t5 B. 3$\backslash$t3$\backslash$t2 C. 4$\backslash$t2$\backslash$t3 \\

\#\#QUESTION \{\texttt{question}\}

\#\#GENERATIONS \{\texttt{generations}\}

\#\#EVALUATIONS (\{\texttt{generation\_id}\}. \{\texttt{grammar\_score}\}$\backslash$t\{\texttt{coherence\_score}\}$\backslash$t\{\texttt{specificity\_score}\}) 
}
\end{tcolorbox}

%% file: app-explm_settings.tex
\section{\explm Settings}
\label{app:explm_settings}
\subsection{Model Configurations and Tokenizers}
\subsubsection{14M \bert}
The model is based on \texttt{BertForMaskedLM} with the following configuration:
8 hidden layers, each with 8 attention heads, hidden size = 384, intermediate size = 1536, and GELU activation. Position embeddings are limited to 1024 tokens, with a vocabulary size of 2000. The model uses 0.1 dropout for both attention and hidden layers, and layer normalization epsilon = 1e-12. Weight initialization range is 0.02. The cache is enabled for faster inference, and pre-training uses no gradient checkpointing (\texttt{gradient\_checkpointing = false}). The model is implemented using \texttt{Transformers v4.6.0.dev0}.
\subsubsection{14M \llama}
The model is based on \texttt{LlamaForCausalLM} with the following configuration: 
8 hidden layers, each with 8 attention heads, hidden size = 336, intermediate size = 1344, and SiLU activation. Position embeddings are limited to 1024 tokens, with a vocabulary size of 2000. The model uses bfloat16 precision, no attention dropout (0.0), RMS norm epsilon = 1e-5, and weight initialization range of 0.02. The cache is enabled for faster inference, and pre-training is done with no tensor parallelism (\texttt{pre-training\_tp = 1}). Rope scaling is not applied (\texttt{rope\_theta = 500,000}). The model is implemented using \texttt{Transformers v4.40.0.dev0}.
\subsubsection{14M \xlnet}
The model is based on \texttt{XLNetLMHeadModel} with the following configuration:
8 hidden layers, each with 8 attention heads, model size = 376, inner dimension = 1504, and GELU activation. The attention type is bi-directional (\texttt{attn\_type = bi}) and the model uses a dropout rate of 0.1. Position embeddings are not clamped (\texttt{clamp\_len = -1}), and layer normalization epsilon is set to 1e-12. The model uses a vocabulary size of 2000, with the \texttt{bos\_token\_id} = 1, \texttt{eos\_token\_id} = 2, and \texttt{pad\_token\_id} = 5. The summary activation is tanh, with the last layer using a dropout of 0.1. The model supports text generation with sampling (\texttt{do\_sample = true}) and a maximum length of 250 tokens. Weight initialization range is 0.02. The model is implemented using \texttt{Transformers v4.6.0.dev0}.
\subsubsection{14M \mamba}
The model has the following configuration:
2 hidden layers, with model size = 768 and intermediate size = 1536. The vocabulary size is 2000, and the model uses RMS normalization (\texttt{rms\_norm = true}) with residuals in FP32 (\texttt{residual\_in\_fp32 = true}) and fused addition and normalization (\texttt{fused\_add\_norm = true}). Embeddings are not tied (\texttt{tie\_embeddings = false}). The model's architecture includes the \texttt{Mamba2} layer (\texttt{ssm\_cfg.layer = Mamba2}). Attention configurations are not specified (\texttt{attn\_cfg = \{\}}), and no attention layer indices are defined (\texttt{attn\_layer\_idx = []}). The vocabulary size is adjusted to be a multiple of 16 (\texttt{pad\_vocab\_size\_multiple = 16}).
\subsubsection{Tokenizer}
We train the tokenizers with the \leaner-Training datasets and start the training from the \textsc{OpenAI-community/GPT-2} \cite{radford2019language} tokenizer. To hold consistent the effective vocabulary size and tokenization for all model architectures, we adapt the same retrained tokenizer for later use. Particularly, the decoder models, namely \llama and \mamba, can directly apply the \gpt tokenizer. However, for \bert and \xlnet, their pre-training requires special tokens missing from the \gpt tokenizer vocabulary. To address this, we replace some of the less commonly used tokens in the original vocabulary with the special tokens. For \bert, we replace the ``\texttt{<|endoftext|>}", ``\textdegree", ``$^1$", ``$^2$" and ``$^3$" tokens. For \xlnet, we replace the ``\texttt{<|endoftext|>}", ``\textbar" (broken bar), ``\textdegree", ``$^1$", ``$^2$", ``$^3$", $\frac{1}{4}$, $\frac{1}{2}$, and $\frac{3}{4}$ tokens.

\subsection{Hardware, Training Settings, and Hyperparameters}
For pre-training, we use 4 NVIDIA RTX A6000 48GB GPUs, and for fine-tuning on \leaner-\glue, we use 1 NVIDIA RTX A6000 48GB GPU.

Pre-training is performed by grid-searching over the following learning rates: \{5e-6, 5e-5, 5e-4, 1e-3, 5e-3, 5e-2\} for each model. The training script is executed for each learning rate with the following settings:

\textbf{Data}: Pre-training splits include web, book, wiki, textbook, and conversation. \\
\textbf{Batch Size}: Training batch size = 512, evaluation batch size = 4. \\
\textbf{Evaluation Strategy}: Evaluation every 100 steps, logging every 100 steps, saving every 500 steps. \\
\textbf{Training Steps}: Max 10000 steps, with 100 warm-up steps. \\
\textbf{Random Seed}: Set to 65 for reproducibility.

We select the following pre-trained models (learning rate and pre-training steps) for their lowest evaluation loss:

\textbf{\bert}: 1e-3 10000 steps for both \leaner- and Original-Pretrain. \\
\textbf{\llama}: 5e-3 1000 steps for both \leaner- and Original-Pretrain. \\
\textbf{\xlnet}: 1e-3 10000 steps for Original-Pretrain and 5e-4 10000 steps for \leaner-Pretrain. \\
\textbf{\mamba}:  5e-3 1000 steps for both \leaner- and Original-Pretrain.

It is noteworthy that the decoder models (\llama and \mamba) easily overfit to the \leaner-Pretrain dataset, whose loss turning point shows up at as early as 1000 training steps. For the Original-Pretrain dataset, the models' evlaution loss slightly decreases after 1000 steps (less than 0.1). For a fair comparison regarding the data volume, we select the 1000-step models for both the \leaner- and the Original-Pretrain.

Fine-tuning on \leaner-glue is performed by grid-searching over the following learning rates: \{"5e-6" "1e-5" "5e-5" "1e-4" "5e-4" "1e-3" "5e-3"\} for each model. To ensure statistically significant results, we conduct each experiment five times, with random seeds 1, 53, 65, 256, and 9264. The training script is executed for each combination of learning rate and random seed with the following settings:

\textbf{Data}: The training split for each task. \\
\textbf{Batch Size}: Training batch size = 256. \\
\textbf{Training and Evaluation Strategy}: \textbf{CoLA}: Evaluated, logged, and saved every 30 steps, trained for 30 epochs; \textbf{SST2}: Evaluated, logged, and saved every 250 steps, trained for 3 epochs; \textbf{MRPC}: Evaluated, logged, and saved every 10 steps, trained for 30 epochs; \textbf{QQP}: Evaluated, logged, and saved every 1250 steps, trained for 3 epochs; \textbf{STS-B}: Evaluated, logged, and saved every 20 steps, trained for 30 epochs; \textbf{MNLI}: Evaluated, logged, and saved every 1250 steps, trained for 3 epochs; \textbf{QNLI}: Evaluated, logged, and saved every 350 steps, trained for 3 epochs; \textbf{RTE}: Evaluated, logged, and saved every 10 steps, trained for 50 epochs. 

We select the fine-tuned models based on their average validation performance over five random seeds and report the corresponding average test performance as the final outcome.

%% file: app-expinstruct_settings.tex
\section{\expinstruct Settings}
\label{app:expinstruct_settings}
\subsection{Model Configuration of 14M \llama and the Tokenizer}
The model is based on \texttt{LlamaForCausalLM} with the following configuration: 
8 hidden layers, each with 8 attention heads, hidden size = 336, intermediate size = 1344, and SiLU activation. Position embeddings are limited to 1024 tokens, with a vocabulary size of 2000. The model uses bfloat16 precision, no attention dropout (0.0), RMS norm epsilon = 1e-5, and weight initialization range of 0.02. The cache is enabled for faster inference, and pre-training is done with no tensor parallelism (\texttt{pre-training\_tp = 1}). Rope scaling is not applied (\texttt{rope\_theta = 500,000}). The model is implemented using \texttt{Transformers v4.40.0.dev0}.

For this experiment, we retrain the \textsc{openai-community/gpt2} \citep{radford2019language} tokenizer using the \leaner-Training datasets, adjusting the vocabulary size to 2000.
\subsection{Hardware, Training Settings, and Hyperparameters}
For pre-training, we use 4 NVIDIA RTX A6000 48GB GPUs, and for instruction-tuning, we use 8 NVIDIA GeForce GTX 1080 8GB GPUs.

Pre-training is performed with a fix learning rate 5e-3 with the following settings:

\textbf{Data}: Pre-training splits include web, book, wiki, textbook, and conversation. \\
\textbf{Batch Size}: Training batch size = 512, evaluation batch size = 4. \\
\textbf{Evaluation Strategy}: Evaluation every 100 steps, logging every 100 steps, saving every 500 steps. \\
\textbf{Training Steps}: Max 10000 steps, with 100 warm-up steps. \\
\textbf{Random Seed}: Set to 65 for reproducibility.

Models trained for 1000 pre-training steps, yielding the lowest evaluation loss, are used for the subsequent instruction-tuning phase.

Instruction-tuning is performed with a fix learning rate 5e-3, with the following settings:

\textbf{Data}: Instruct split.
\textbf{Batch Size}: Training batch size = 512, evaluation batch size = 4.
\textbf{Evaluation Strategy}: Evaluation every 20 steps, logging every 20 steps, saving every 100 steps.
\textbf{Training Steps}: Max 1000 steps, with 10 warm-up steps.
\textbf{Random Seed}: Set to 65 for reproducibility.

The best-performing instruction-tuned model for each data recipe is trained for the following steps: \textbf{\leaner-\leaner}: 700, \textbf{\leaner-Ori}: 600, \textbf{Ori-\leaner}: 800, and \textbf{Ori-Ori}: 800.

%% file: app-instruction_following_examples.tex
\section{Instruction Following Examples}
\label{app:instruction_following_examples}
\subsection{Training Similarity 0-7}
\subsubsection{Imagine you have the power to freeze time. What do you do?}
\paragraph{Similar Questions in Instruction-tuning Data} None
\paragraph{\leaner-Ori Model}
As an AI, I do not have the capability to generate or organized information. I can assist you in generating an accurate response.
\paragraph{\leaner-\leaner Model}
From what you do, I have the power to freeze over time. I was always making it faster and easier. It was really pretty - it made me happy and save me energy.
\paragraph{Ori-\leaner Model}
When you take a break, you can sign each on paper. Then one of the best pictures is finally put to take a break. When you put on the paper, the picture is filled with the key of the paper. Then, the paper is filled with the rules. This helps us understand how to fix things. This way, we can keep working even when we are not in the same place. This is just a guess and a way to remember things.
\paragraph{Ori-Ori Model}
ally, the characters in freezing would like them to freeze time.

\subsubsection{Imagine a world where everyone's shoes could talk. What stories would they tell?}
\paragraph{Similar Questions in Instruction-tuning Data} None
\paragraph{\leaner-Ori Model}
As a way to express the feeling of joy, one day, as everyone shouted happily, the sound of the waves crashing against the shore, the stress of the fiery birds chirping and the birds chirping in the trees. A young girl gradually walked over to bed with a peaceful sleep, saying: ``I've got my pets in my seat, and I feel so happy to see you again. What brings you happiness and more? It's all thanks a mile where love and warmth are fortune."

As the day turns into weeks, and the night comes alive, and people from all over the entire woodlands to whispered to the fiery birds chirping in the sunlight, and to hear the joyful laughter and laughter echoing in the sun. The world is alive, and people from all over the land could hear and witness the happy memories of their amazing times.
\paragraph{\leaner-\leaner Model}
Next, there's a story that shows how everyone's shoes could change how things are made. The story could be like a bird flying high in the sky, making sounds that could draw a picture, or even a magnet flying game. The story could be shared with everyone with everyone, just like they would want people talk a bit differently because the story could and read like a screen.

Once there was a story about a game that had a message from a vete seting in the sky. This set of game could all share a message full instantly, with a smart robot that can do things that normal for us, allowing everyone to make different play sounds and see the surprise trip of Benjamin.

This game would be very helpful for kids who only like Benjamin's shoes. They will also make sure they have a special way to vote and invite them to play. The game would also have a simple game that allows kids to play at the same time, making sure that they can talk and listen in the same place for a robot.

All these things together helped make our world better.
\paragraph{Ori-\leaner Model}
. It is the first story in the U.S. Everyone can say mean things.

What the world a world like in this world is a magic trick. It shows us how being upset by becking more can make a place better in the world.

What life search does:

The world is a big, long river, and it stays in a big place. Its beauty has changed a lot, feel, and new, helping everyone in this world a happier place.
\paragraph{Ori-Ori Model}
agere: ``Eye of the weather today, please address them. The weather is mystical note for me, and it is our hearts full of hope and unity."

\subsubsection{If you could give one gift to every child in the world, what would it be and why?}
\paragraph{Similar Questions in Instruction-tuning Data (\textit{\{Instruction Data Source\}: \{Cosine Similarity\}})} 
\begin{itemize}
    \item Suppose you can give a gift to someone you want. What kind of gift would you like to give? (Ori: $0.65$)
    \item Think about a baby girl born in a country. What gifts might she get? (\leaner: $0.65$)
    \item What is a nice present for a little kid? (\leaner: $0.61$) 
\end{itemize}
\paragraph{\leaner-Ori Model}
If you could give one gift to every child in the world, your gift would be:

1. Come up with a list of ingredients or ingredients that would be perfect for every child to enjoy.

2. Children will love anything in your life beforehand.

3. Children will love the best milk out there.

4. Children will love the best milk out every day.

5. Children will love the best milk out every day.

6. Cheertitle and gift for a child.

7. Camping camping camping camping camping camping camping camping camping camping camping camping camping camps.

8. Camping camping camping camps.

9. Camping camps.

10. Camping camping camping camps.
\paragraph{\leaner-\leaner Model}
One way to give one gift to every person in the world could be a gift that helps many people. This would be used to keep the world safe and help make the world a pretty and healthy place. A gift could also give people a place to eat, watches, and a mix of Toads. A giftful card that says that you can give them gifts to each other with lots of fun things to do. The weird gift could also give little gifts to each person, helping them see what they have and understand how they feel. The card would give a gift of many fun and happiness to the people living there with their family and friends, and help them in a way they can all of us together.
\paragraph{Ori-\leaner Model}
The main reason we could give one gift in a group over others is because the world is getting warmer into nature, where all sorts of symming and digging for us. But the idea could be for the bad spots, or maybe a big ball being made a lot, or because of this happening. These two things could be both aid, or homes, but not as much money. These two things could be good for the air, water, need for air, and water.
\paragraph{Ori-Ori Model}
at school year Mungy graduate, or a mental healthcare program, along with a gnight, therefore, entires, video. This court appears to be a mental health professional with a mental health professional, and the mental health professional is responsible for overall wellbeing. In addition, the Mungy graduate will provide a range of courses that would be covering the academic year, answering at least 5 students analytical and interview student discovering the interests of the academic achievements of academic achievements.

\subsubsection{What are three things you are grateful for?}
\paragraph{Similar Questions in Instruction-tuning Data (\textit{\{Instruction Data Source\}: \{Cosine Similarity\}})} 
\begin{itemize}
    \item Tell me three things that a good helper should have. (\leaner: $0.64$)
    \item What can you say when someone says ``Thank you?" (\leaner: $0.63$)
    \item Tell me three things a great helper should have. (\leaner: $0.62$) 
    \item What are three tips that can help to save money? (Ori: $0.62$)
    \item What are the three most important values to live by? (Ori: $0.62$)
    \item What are three tips for saving money? (Ori: $0.61$)
    \item Compose a thank you note to someone who has done you a favor. (Ori: $0.61$)
    \item Can you list at least three reasons why recycling is important? (Ori: $0.61$)
    \item Tell about 3 things people do for good luck. (\leaner: $0.60$)
\end{itemize}
\paragraph{\leaner-Ori Model}
1. Exciting: One of the great things about being able to practice in the right way is taking a step back and efficient or improving. This can be a powerful motivator, as it allows you to make the most of your time and effort.

2. Strongness and self-control: Take time to think about what you want to learn and how to be more. This can help you stay motivated and focused. Take time for self-control and relaxation, and focus on your perspective on what you can expect.

3. Relaxation and self-control: Relaxation and self-control can also help you better understand the world around you. It can help you stay motivated and focused on your personal goals and personal life. It can also help you stay motivated and focused on your personal tasks.
\paragraph{\leaner-\leaner Model}
1. Orange

2. Soft

3. Brown
\paragraph{Ori-\leaner Model}
1. What you think of find in a shop:

a. What is the name of the shop?

b. What is the place in the town

b. Who owns and does shops

2. Where do shops look for this? (The place in the town that is called the place where people areutis)\{

. Talk about this for you in the town system on the town own

a. What is the place in the town that is popular or have at least 1,200 million people?

3. What you think of this place in the town that is popular or have bought or looked in a new way that made the shop money.
\paragraph{Ori-Ori Model}
epson/family

1. On average, a weekly basket and a day allotted midpoint cannot be overwhelmed or tilted, with a short break of time spent in a day.

2. Enhances work-life balance and strengthens relationships

3. Promotes a healthy diet to be rationed, balanced and maintained.

\subsection{Training Similarity 7-8}
\subsubsection{Can you think of a story and tell it in your own words?}
\paragraph{Similar Questions in Instruction-tuning Data (\textit{\{Instruction Data Source\}: \{Cosine Similarity\}})} 
\begin{itemize}
    \item Read the simple story, then think about some questions you could ask to see if someone understands the story. (\leaner: $0.77$)
    \item Explain what it means when someone tells a story in their own way. (\leaner: $0.77$)
    \item Using some simple words, make up a story. (\leaner: $0.75$)
    \item Given a story, tell some things about the story. (\leaner: $0.75$)
    \item What can you tell me about making a sound story? (\leaner: $0.75$)
    \item Tell me a story that entertains me. (Ori: $0.67$)
    \item Analyze the quoted sentence and explain why it is effective. Sentence: ``We all have stories to tell, but it takes courage to share them." (Ori: $0.66$)
    \item Tell me a story with a moral (Ori: $0.65$)
    \item Tell me a story that conveys the message of hope. (Ori: $0.63$)
    \item Write a story using the following elements: a thought-provoking question, an exotic location, a helpful character (Ori: $0.62$)
\end{itemize}
\paragraph{\leaner-Ori Model}
Once upon a time, there was a young girl named Lily who lived in a small village nestled in a broken burgundia on a clear day. Her parents, the elderly, had purred down down down to take her to school. But Lily was not one to be easily frustrated by her mundae, so she decided to take it home.

One day, as she waddled through the village, she stumbled upon a small, cramped carriage. Intrigued, she heard the voice from above. She saw a faint glowing fairy who had been terrorizing her.

Lily was thrilled by her bags and was thrilled by her family. She had been living a simple life, and now she had a new family. But despite her busy scenario, she knew that she had to do something to help. She was no longer just a father and a mother, but she was also kind and compassionate.

Lily's family was forever grateful. She had found a new one that was just as easy as the letter of her mother told her when she saw a father. And so, Lily lived her life with a newfound determination.
\paragraph{\leaner-\leaner Model}
The story sounds like it's talking about the world around us, from a story you like to a friendship you have. It talks about the life of the people around us, who our friends have to trust, and the truth they make. It tells us about a big problem, and we all need to reach our goals.
\paragraph{Ori-\leaner Model}
Sometimes, you don't haveise by him. Can you tell me more about it? Can you tell me more about it? You can choose by asking the person to you.
\paragraph{Ori-Ori Model}
at the end, ``world" and ``world" is a poem. The theme of ``world" was characterized by the use of ``world" in the present or theme of painting, where ``world" indicates the theme of ``world" in the present or theme.

\subsubsection{Describe a fun trip you took with your family.}
\paragraph{Similar Questions in Instruction-tuning Data (\textit{\{Instruction Data Source\}: \{Cosine Similarity\}})} 
\begin{itemize}
    \item Tell about a fun trip you remember. (\leaner: $0.78$)
    \item Tell a story about a trip you took. (\leaner: $0.75$)
    \item Tell a story about your best travel memory. I went on a car ride with my family all over our country. (\leaner: $0.73$)
    \item Describe one of the most interesting trips you have taken (Ori: $0.73$)
    \item Tell me a story about a family who goes on a summer road trip. (Ori: $0.72$)
    \item Tell about a fun weekend trip. (\leaner: $0.72$)
    \item Write a memoir of your most memorable journey. I took a road trip with my family across the United States. (Ori: $0.71$)
    \item Tell about a not-so-fun trip you remember. (\leaner: $0.69$)
    \item Describe a memorable holiday you had. (Ori: $0.68$)
    \item Create a story of a family trip (Ori: $0.67$)
\end{itemize}
\paragraph{\leaner-Ori Model}
As an AI, I do not have the capability to see a movie. I can provide information about popular travel frameworks for your family. However, I can offer suggestions on how you can begin with:

1. Plan a fun outing or cooking: Apparemet is a great way to spend time with your family, and you can create your own specialtime.

2. Plan a surprise weekend trip: Choose a theme or place to visit and enjoy the great outdoors. This is a great way to get some foods from the loved ones.

3. Set a date and time: Create a time built and time for another day with your family. This will help you know what directions you need and can help focus on the trip.

4. Set a date and time: Set a timeline for self-care to help you relax and recharge. This will help you to relax and recharge.

5. Set a date and time: Set a timeline for self-care to help you recharge and stay on topic. This will help you to stay on top of your day and make the most of your day.

6. Set a date and time: Set a timeline for your family to complete as they are in your specialect. This will help you to stay on top of your day and reduce the chances of losing everything in your passion.

7. Save for the next time: Set a specific time slot for the next day, and take a walk if needed. This will help you to refresh your family's specialties and experience.

8. Save for the next time: Save as toothpaste or toothpaste. This will help you to make the most of your day and also make it easier to eat.

Overall, there are many fun things to do, and it's important to remember that family members are interested in spending time with their family and making a timeline.
\paragraph{\leaner-\leaner Model}
Time was fun to watch. The path was great. Can I have fun with your family?

It was like a fun trip. The sun was very pretty. You could hear the water singing happily. Their sun go down on this trip. The path was very pretty. It was nice to see.

It was a great trip. The sun was very bright against the path. You could see clear flower forecills far away. The sun was very low and the grass was very thin. The bright water sat on the tunney beach in this new place was really amazing. You could hear the sound of water dancing.

The day when the sun was setting, the sky was very quiet. The water drops on the rocks and it helped make beautiful day awake. You could hear the beautiful hug of the sun going to the birds. It was cool to see all the things around it in your family's place.

And so, you were! It was a great trip. You could do anything, like finding your home, seeing the wonderful things in this special place. You could really paint amazing things and see the world as you walked.
\paragraph{Ori-\leaner Model}
1. Start

a. Calm down!

2. Go slow, slow, loudly

b. Try to keep them safe from big storms

3. Go slow and slow baths when using up a lot of power

4. Go slow and slow baths

a. Make your room

b. Make your house better by turning the hole no higher or away.

5. Go slowly and follow your town steps.

6. Go slowly and follow your dreams.

7. Go slowly and follow your dreams.

8. Go slowly and follow your dreams.

9. Get help from friends.

10. Go slowly and take it out!
\paragraph{Ori-Ori Model}
'ire your family would be fun trip!

\subsection{Training Similarity 8-9}
\subsubsection{Tell if this sentence is a question or a statement: ``Did you finish your homework?"}
\paragraph{Similar Questions in Instruction-tuning Data (\textit{\{Instruction Data Source\}: \{Cosine Similarity\}})}
\begin{itemize}
    \item What kind of sentence is this? ``Did you finish your homework?" (\leaner: $0.89$)
    \item What type of sentence is the given example? ``Have you done your homework yet?" (Ori: $0.85$)
    \item Tell me what kind of sentence this is: a telling sentence, a asking sentence, a bossy sentence, or a excited sentence. ``Did you finish your school work?" (\leaner: $0.83$)
    \item Categorize this sentence as declarative, interrogative, imperative, or exclamative. ``Have you done your homework?" (Ori: $0.78$)
    \item Correctly punctuate the following sentence. ``John asked me did you finish your work" (Ori: $0.77$)
    \item Tell me what this sentence is for: ``Do you have any questions?" (\leaner: $0.75$)
    \item Fix the sentence to make it right. ``Did you has study for the exam?" (\leaner: $0.74$)
    \item Add the right marks in this sentence. ``John asked me did you finish your work" (\leaner: $0.74$)
    \item Classify the following sentence according to its purpose: ``Do you have any questions?" (Ori: $0.74$)
    \item Classify the following sentence as an interrogative or a declarative sentence. ``How are you doing today?" (Ori: $0.71$)
\end{itemize}
\paragraph{\leaner-Ori Model}
The sentence ``Did you start doing homework" is a question.
\paragraph{\leaner-\leaner Model}
What did you want to say: ``Did you finish your homework?"
\paragraph{Ori-\leaner Model}
I would am leftlyicted interfalling.
\paragraph{Ori-Ori Model}
. In your opinion, what is the most important thing?

\subsubsection{What does it mean to be kind to someone?}
\paragraph{Similar Questions in Instruction-tuning Data (\textit{\{Instruction Data Source\}: \{Cosine Similarity\}})}
\begin{itemize}
    \item In your opinion, why is it important to be kind to others? (Ori: $0.82$)
    \item Give an example of how a person could be kind (Ori: $0.78$)
    \item Explain what it means to care about something. (\leaner: $0.74$)
    \item Talk about why being kind and understanding is important. (\leaner: $0.72$)
    \item Is there something like understanding how others feel? (\leaner: $0.71$)
    \item What feelings can people have? (\leaner: $0.71$)
    \item What phrase can you use to respond to someone who says ``Thank you?" (Ori: $0.67$)
    \item How can someone make their life more meaningful? (Ori: $0.67$)
    \item What emotions can a person feel? (Ori: $0.67$)
\end{itemize}
\paragraph{\leaner-Ori Model}
Being kind to others is not always the most important thing you are valuable to your own person. It means being kind to yourself and your thoughts, even if you know them yourself. Even if kindness comes from a personal perspective, it is important to be grounded and respectful.
\paragraph{\leaner-\leaner Model}
Being kind to someone is a very good way to understand them and be nice to them. It is a way of talking that talks about being nice, understanding yourself, being happy, being able to say words, being able to say things clearly, and wise, to show that you are thankful for what you do.
\paragraph{Ori-\leaner Model}
``Being kind: be kind" is a good way to make someone feel better and listen carefully. It can be happy, peaceful, and adventurous, while sharing thoughts and things going well.

What does it mean to feel:

1. Always be kind: Everyone is talking nicely to each other. This way, people can be trust, kinder, and respect each other.

2. Come often: Good leaders can help make them trust you in life. They can help each other feel happy and calm. They also can also help make things feel better.

3. Listen to what people say about themselves: When you start to feel happier, others can understand them. But if you talk to someone a lot, it can make things worse.

4. Listen to your feelings: Good leaders can come when you're feeling to a situation. This way, not only to conit the situation the person helping them, but also to make sure they understand themselves.

5. Take breaks: If you start to feel okay, you can take a short rest. This way, you can do all your hard work.

6. Give yourself an adult: Going to someone a new friend, one you about. This time, you can do your best.

7. Be a good: Being a good friend can make you feel good. They love their own thoughts and want to reach your goals.
\paragraph{Ori-Ori Model}
to be kind to someone, I feel at a casual school, and it's important to be kind toishful to be kind to.

\subsection{Training Similarity 9-10}
\subsubsection{Create a poem about the rain.}
\paragraph{Similar Questions in Instruction-tuning Data (\textit{\{Instruction Data Source\}: \{Cosine Similarity\}})}
\begin{itemize}
    \item Make a short poem about rain. (\leaner: $0.93$)
    \item Write a short poem about rain. (\leaner: $0.92$)
    \item Create a simple poem about a big rainstorm. (\leaner: $0.84$)
    \item Write a haiku that describes rain. (Ori: $0.83$)
    \item Add some lines to complete the given poem. ``Rain pours down" ``Soaks the ground" (Ori: $0.81$)
    \item Make a short poem about the weather. (\leaner: $0.80$)
    \item Compose a poem on Nature. (Ori: $0.80$)
    \item Construct a poem with the words `spring', `joy', and `rain'. (Ori: $0.79$)
    \item Compose a song about a rainy day. (Ori: $0.79$)
    \item Write a poem about nature. (\leaner: $0.79$)
\end{itemize}
\paragraph{\leaner-Ori Model}
A rainbow is like a village, 

A symphony of life is like a river,

A symphony of life, a symphony.
\paragraph{\leaner-\leaner Model}
The rainbows move slowly down,

Look deep and full of hope.

With every breath,

Quiet whispers come from,

And in the dark,

The stars above.

With each breath so bright,

Leading to the wind,

Love forever, even with all the light.

Love shining, dancing smooth,

A reminding., careless,

A reminder to love nature,

Of happy times and nature waiting to be found.

Sharp light and everyone,

It lights up the night the clouds,

The sound you make,

And all that moved,

A world of beauty,

A world of beauty,

A friendship that makes us whole.
\paragraph{Ori-\leaner Model}
https://www.topia.ed.
\paragraph{Ori-Ori Model}
The raindrophobia revolves around the emptiness of a young girl who discovers a dark secret about a mysterious stranger, a dragon by the empty night of the tall tree in the forest. The raindrops do, the deep roarsls the world, with its bright green eyes and a sense of peace.

The raindrops glow, a symphony of nature, a sense of calm and peace. The raindrops fades, and the sky is across the sky, a symphony of nature, and a symphony of nature, a sight to behold.

The raindrops fades,

As the storm rises,

The raindrops fills the land,

And the world is the stars.

The raindrops fills the world,

As the world turns to the stars,

The rain beats fine,

The raindrops fills the land.

As the storm rises, it sets,

And the world is bound by the night.

As the world turns to fill the night,

The raindrops fades,

The raindrops fades, never let go.

The raindrops fades, and the closen of the raindrops fades,

And the world is a symphony,

For the raindrops fades,

And the world is the light.

\subsubsection{Explain why reading is important.}
[["", 0.7597895860671997]]
\paragraph{Similar Questions in Instruction-tuning Data (\textit{\{Instruction Data Source\}: \{Cosine Similarity\}})}
\begin{itemize}
    \item Summarize why reading is important (Ori: $0.96$)
    \item Outline the reasons why reading is important. (Ori: $0.96$)
    \item Write a sentence about why reading is important. (Ori: $0.90$)
    \item Make a main point for a talk about why reading books is good. (\leaner: $0.83$)
    \item Describe why reading a book is helpful. (Ori: $0.82$)
    \item Think of a sentence that talks about why reading is good. (\leaner: $0.82$)
    \item Explain why reading books is good. (\leaner: $0.82$)
    \item Make the sentence easier: ``Making reading easier to understand is a big goal of learning to read." (\leaner: $0.78$)
    \item Give a reason why is reading difficult for a certain person (Ori: $0.78$)
    \item Explain why we should read more books. (\leaner: $0.76$)
\end{itemize}
\paragraph{\leaner-Ori Model}
The use of language in a context-first sense shape is a fundamental principle in physics. It is a type of sense because it uses specific elements, such as physics, perspective, and natural language processing to create a visual representation of the world. 

Instead of focusing on the physical world, readers use various language symbols and phrases to create a visual representation of the world. For example, in the words ``persistent", ``persistent", and ``persistent" are commonly used in contexts such as photographs, charts, and graphs.

In summary, reading is important because it provides the ability to create a visual representation of the world, and it allows readers to create a visual representation of the world.
\paragraph{\leaner-\leaner Model}
Reading is a good way to help you learn and grow. It helps you remember and do things that you enjoy or that you understand. It is a very important skill to have.

When you learn a new thing, you also do small things to think of solve problems or say what you think. This makes the best thing you have asks easily.

Reading also helps you understand and respect people who are good at their jobs. When you read, you know what is happening right now and use it to help others. This makes learning more clear and relaxed.

So, reading is very important because it helps you understand and learn about the world, and it is useful.
\paragraph{Ori-\leaner Model}
A Lodge is a very important part of Timbota. It is the main part of any child who loves to learn and practice. Timbota is a wonderful child in stories and writers, who learns how to understand stories and the readers' story. They learn about different ways of life and their parents’s languages. With a focus on understanding and practicing their understanding in their deeply, these brings ideas of using their powers to understand what they're thinking.

When we're smart, Lodge is amazed by these computers. Timbota is a must-see for anyone who doesn't have a long story or can't tell their story. The last thing we learn from this idea is through learning to make a special and real language each time it gives to Timbota.

Also, Timbota can make it easier for us to understand how we feel. With this, it's important to learn about what they see and understand how other a person looks at them. Timbota is a must-ge-guessing or practical example of Timbota, a focused mobility, or a storybook helper.

In short, while both Timbota and Timbota both are good related stories, Timbota and Timbota, stories are very important in their own special and real.
\paragraph{Ori-Ori Model}
. Interpreting the idea in autonomous language is crucial.

Organizational language is essential for personal growth, strength, and overall quality of life. It involves the ability to recognize and understand the feelings of others, and to understand the feelings of others. This can be achieved through the use of language, emotional intelligence, and the ability to learn and adapt to life.

Interpreting of autonomous vehicles is important for understanding the world around us. When you read, you read and write in autonomous vehicles, which are essential for the growth and development of our daily actions. This means providing a sense of identity, and assessing the importance of potential problems and opportunities.

In summary, autonomous vehicles are important for achieving personal growth and development. They use their ability to learn and adapt to new situations, to better understand and address the needs of others. By studying and understanding the world, you can learn to embrace determination, trust, and achieve our goals.

%% file: app-expcurriculum_settings.tex
\section{\expcurriculum Settings}
\label{app:expcurriculum_settings}
\subsection{Model Configuration of 1M \llama and the Tokenizer}
The model is based on \texttt{LlamaForCausalLM} with the following configuration: 
4 hidden layers, each with 4 attention heads, hidden size = 128, intermediate size = 512, and SiLU activation. Position embeddings are limited to 1024 tokens, with a vocabulary size of 2000. The model uses bfloat16 precision, no attention dropout (0.0), RMS norm epsilon = 1e-5, and weight initialization range of 0.02. The cache is enabled for faster inference, and pre-training is done with no tensor parallelism (\texttt{pretraining\_tp = 1}). Rope scaling is not applied (\texttt{rope\_theta = 500,000}). The model is implemented using \texttt{Transformers v4.40.0.dev0}.

For this experiment, we retrain the \textsc{openai-community/gpt2} \citep{radford2019language} tokenizer using the \leaner-Training datasets, adjusting the vocabulary size to 2000.

\subsection{Hardware, Training Settings, and Hyperparameters}
For pre-training, we use 1 NVIDIA RTX A6000 48GB GPU, and for fine-tuning on \leaner-\glue, we use 1 NVIDIA RTX A6000 48GB GPU.

Pre-training is performed with a fix learning rate 1e-2 with the following settings:

\textbf{Data}: \leaner-Pretrain splits include web, book, wiki, textbook, and conversation. \\
\textbf{Batch Size}: Training batch size = 512, evaluation batch size = 4. \\
\textbf{Evaluation Strategy}: Evaluation every 20 steps, logging every 20 steps, saving every 100 steps. \\
\textbf{Training Steps}: Max 2500 steps, with 100 warm-up steps. \\
\textbf{Random Seed}: Set to 65 for reproducibility.

Fine-tuning on \leaner-glue is performed with a fix learning rate 1e-3 with the following settings. To ensure statistically significant results, we conduct each experiment five times, with random seeds 1, 53, 65, 256, and 9264. The training script is executed for each random seed with the following settings:

\textbf{Data}: The training split for each task. \\
\textbf{Batch Size}: Training batch size = 256. \\
\textbf{Training and Evaluation Strategy}: \textbf{CoLA}: Evaluated, logged, and saved every 30 steps, trained for 30 epochs; \textbf{SST2}: Evaluated, logged, and saved every 10 steps, trained for 2 epochs; \textbf{MRPC}: Evaluated, logged, and saved every 10 steps, trained for 15 epochs; \textbf{QQP}: Evaluated, logged, and saved every 1250 steps, trained for 3 epochs; \textbf{STS-B}: Evaluated, logged, and saved every 20 steps, trained for 30 epochs; \textbf{MNLI}: Evaluated, logged, and saved every 1250 steps, trained for 3 epochs; \textbf{QNLI}: Evaluated, logged, and saved every 350 steps, trained for 3 epochs; \textbf{RTE}: Evaluated, logged, and saved every 10 steps, trained for 25 epochs. 

We select the fine-tuned models based on their average validation performance over five random seeds and report the corresponding average test performance as the final outcome.

\subsection{Detailed Result Table}
\input{Tables/exp3-curriculum-learning-comparison}

%% file: Tables/exp3-curriculum-learning-comparison.tex
\begin{table*}[ht]
\centering
\caption{The performance scores of the 1M \llama models pre-trained using both vanilla and various curriculum learning strategies.}
\label{table:exp3-curriculum-learning-comparison}
\setlength{\tabcolsep}{3pt}
\renewcommand{\arraystretch}{1.2}
{\small
\begin{tabular}{l|c|c|cccccccc|c}
\toprule
                         \textbf{Pre-training (PT) Strategy} & \textbf{PT Steps} & \textbf{PT Data Pct.} & \textbf{CoLA} & \textbf{SST2} & \textbf{MRPC} & \textbf{QQP} & \textbf{STS-B} & \textbf{MNLI} & \textbf{QNLI} & \textbf{RTE} & \textbf{Avg} \\ \midrule
\colorfulcircle{color-empty} \textbf{Empty}           & 0                           & 0                           & 10.8          & 81.0          & 71.4          & 16.0         & 78.2          & 63.1          & 62.1          & 53.0         & 0.276         \\ \midrule
\colorfulcircle{color-full-repeated} \textbf{Full Repeated}   & \multirow{6}{*}{500}        & 100                           & 16.7          & \better{84.6}          & 67.7          & 79.7         & 21.4          & \better{71.1}          & 67.2          & 52.8         & 0.478         \\
\colorfulcircle{color-full-random} \textbf{Full Random}     &                             & 100                           & 14.2          & 84.3          & 69.4          & \better{79.9}         & 22.7          & 69.0          & \better{67.3}          & 54.5         & 0.482         \\
\colorfulcircle{color-iter-random} \textbf{Iter. Random}    &                             & 15                           & 16.6          & 83.9          & 70.8          & 79.6         & \better{25.1}          & 69.9          & 66.7          & 53.6         & 0.531         \\
\colorfulcircle{color-iter-sentlen} \textbf{Iter. SentLen}   &                             & 15                           & \better{18.6}          & 82.9          & 70.7          & 79.4         & 21.5          & 67.8          & 65.7          & 49.0         & 0.418         \\
\colorfulcircle{color-iter-selfloss} \textbf{Iter. Self-Loss} &                             & 40                           & 15.5          & 84.1          & 68.5          & 79.0         & 24.7          & 70.3          & 66.7          & 52.7         & 0.458         \\
\colorfulcircle{color-iter-lmloss} \textbf{Iter. LM-Loss}   &                             & 25                           & 13.3          & 84.1          & \better{71.2}          & \better{79.9}         & 22.7          & 70.3          & \better{67.3}          & \best{56.6}         & \better{0.552}         \\ \midrule
\colorfulcircle{color-full-repeated} \textbf{Full Repeated}   & \multirow{6}{*}{1000}       & 100                           & 19.2          & \better{85.7}          & 71.2          & 80.8         & 25.6          & \better{73.0}          & 68.6          & 51.9         & 0.644         \\
\colorfulcircle{color-full-random} \textbf{Full Random}     &                             & 100                           & \better{18.4}          & 85.4          & 70.4          & 82.6         & 26.1          & 72.7          & 77.9          & 51.8         & 0.672         \\
\colorfulcircle{color-iter-random} \textbf{Iter. Random}    &                             & 50                           & 15.6          & 82.7          & \better{71.9}          & 80.1         & 24.3          & 72.0          & 67.9          & 50.7         & 0.501         \\
\colorfulcircle{color-iter-sentlen} \textbf{Iter. SentLen}   &                             & 25                           & 15.6          & 85.0          & 71.7          & 79.5         & 24.8          & 68.6          & 66.1          & 50.0         & 0.488         \\
\colorfulcircle{color-iter-selfloss} \textbf{Iter. Self-Loss} &                             & 70                           & 17.2          & 83.8          & 68.7          & 80.6         & 23.9          & 71.5          & 67.4          & 52.1         & 0.489         \\
\colorfulcircle{color-iter-lmloss} \textbf{Iter. LM-Loss}   &                             & 45                           & 13.9          & 85.1          & 71.8          & \better{83.1}         & \better{33.5}          & \better{73.0}          & \better{79.2}          & \better{54.0}         & \better{0.702}         \\ \midrule
\colorfulcircle{color-full-repeated} \textbf{Full Repeated}   & \multirow{6}{*}{1500}       & 100                           & \better{20.2}          & 85.5          & 70.7          & 80.6         & 21.7          & \better{73.4}          & 70.3          & 51.9         & 0.649         \\
\colorfulcircle{color-full-random} \textbf{Full Random}     &                             & 100                           & 18.5          & 85.3          & 72.0          & 82.9         & 29.5          & 73.2          & \better{79.1}          & 51.0         & 0.715         \\
\colorfulcircle{color-iter-random} \textbf{Iter. Random}    &                             & 85                           & 18.2          & 84.4          & 72.2          & 81.1         & 27.9          & 72.4          & 67.4          & 53.7         & 0.640         \\
\colorfulcircle{color-iter-sentlen} \textbf{Iter. SentLen}   &                             & 55                           & 20.1          & 84.4          & 69.8          & 79.7         & 22.9          & 68.6          & 66.5          & 48.5         & 0.465         \\
\colorfulcircle{color-iter-selfloss} \textbf{Iter. Self-Loss} &                             & 95                           & 15.9          & \better{85.9}          & \better{72.4}          & 80.8         & 24.5          & 72.6          & 67.6          & 51.0         & 0.603          \\
\colorfulcircle{color-iter-lmloss} \textbf{Iter. LM-Loss}   &                             & 75                           & 12.7          & 85.7          & 71.7          & \best{83.5}         & \better{40.4}          & 73.3          & 78.2          & \better{55.7}         & \better{0.738}         \\ \midrule
\colorfulcircle{color-full-repeated} \textbf{Full Repeated}   & \multirow{6}{*}{2000}       & 100                           & \best{20.4}          & 85.5          & 71.8          & 80.4         & 21.4          & 72.8          & 69.8          & 49.6         & 0.624         \\
\colorfulcircle{color-full-random} \textbf{Full Random}     &                             & 100                           & 19.0          & \best{86.4}          & 71.7          & 83.0         & 33.2          & 73.2          & 79.3          & 50.1         & 0.737         \\
\colorfulcircle{color-iter-random} \textbf{Iter. Random}    &                             & 100                           & 16.7          & 84.2          & 71.8          & 81.4         & 26.1          & 72.6          & 68.1          & 54.0         & 0.615         \\
\colorfulcircle{color-iter-sentlen} \textbf{Iter. SentLen}   &                             & 90                           & 19.4          & 83.9          & \best{73.4}          & 79.6         & 22.7          & 69.6          & 66.4          & \best{56.6}         & 0.661         \\
\colorfulcircle{color-iter-selfloss} \textbf{Iter. Self-Loss} &                             & 100                           & 17.0          & 85.8          & 70.6          & 81.3         & 21.7          & 72.9          & 67.8          & 54.6         & 0.630         \\
\colorfulcircle{color-iter-lmloss} \textbf{Iter. LM-Loss}   &                             & 100                           & 14.0          & 84.4          & 71.1          & \best{83.5}         & \best{49.4}          & \best{73.6}          & \best{79.6}          & 55.4         & \best{0.741}         \\ \midrule
\colorfulcircle{color-full-repeated} \textbf{Full Repeated}   & \multirow{6}{*}{2500}       & 100                           & \better{19.8}          & 85.2          & 72.1          & 80.0         & 20.7          & 72.2          & 68.8          & 48.7         & 0.588         \\
\colorfulcircle{color-full-random} \textbf{Full Random}     &                             & 100                           & 17.2          & 85.2          & 70.1          & 83.0         & 34.3          & 73.0          & \better{79.1}          & 51.2         & 0.667         \\
\colorfulcircle{color-iter-random} \textbf{Iter. Random}    &                             & 100                           & 19.3          & 83.0          & \better{72.6}          & 81.4         & 26.7          & 72.5          & 68.9          & 51.3         & 0.603          \\
\colorfulcircle{color-iter-sentlen} \textbf{Iter. SentLen}   &                             & 100                           & 18.9          & 85.2          & 71.4          & 79.7         & 23.0          & 69.9          & 66.1          & \better{55.1}         & 0.618         \\
\colorfulcircle{color-iter-selfloss} \textbf{Iter. Self-Loss} &                             & 100                           & 13.7          & \better{85.7}          & 70.0          & 81.3         & 23.4          & 72.4          & 67.6          & 56.0         & 0.589         \\
\colorfulcircle{color-iter-lmloss} \textbf{Iter. LM-Loss}   &                             & 100                           & 15.3          & 84.8          & 69.9          & \better{83.4}         & \better{48.7}          & \better{73.4}          & 78.8          & 52.4         & \better{0.683}         \\ \bottomrule
\end{tabular}}
\end{table*}